\documentclass[11pt,a4paper]{article}
\usepackage{times,latexsym}
\usepackage{url}
\usepackage[T1]{fontenc}

\usepackage[acceptedWithA]{tacl2021v1}
\usepackage{xspace,mfirstuc,tabulary}

\newif\iftaclinstructions
\taclinstructionsfalse 
\iftaclinstructions
\renewcommand{\confidential}{}
\renewcommand{\anonsubtext}{(No author info supplied here, for consistency with
TACL-submission anonymization requirements)}
\newcommand{\instr}
\fi

\iftaclpubformat 

\else

\fi

\usepackage{graphicx,bm}
\usepackage{caption}
\usepackage{subcaption}
\usepackage{amsmath,amsfonts}
\usepackage{varwidth}
\usepackage{rotating}
\usepackage{multirow}
\usepackage{booktabs}
\usepackage{makecell}
\usepackage{colortbl}
\usepackage{vcell}
\usepackage[export]{adjustbox}
\usepackage[ruled]{algorithm}
\usepackage[endLComment={},rightComments=false]{algpseudocodex}
\algrenewcommand\algorithmicrequire{\textbf{Input:}}
\algrenewcommand\algorithmicensure{\textbf{Output:}}
\usepackage{soul}
\setul{2pt}{.4pt}
\newcommand{\uline}[1]{\ul{#1}}
\renewcommand{\underline}[1]{\ul{#1}}
\usepackage{pifont}
\usepackage{tabularx}
\newcolumntype{R}{>{\raggedleft\arraybackslash}X}
\usepackage{stfloats}
\usepackage{enumitem}
\usepackage{nicefrac}
\usepackage[htt]{hyphenat}

\newcommand{\mysub}[2]{{#1}_{{}_#2}}

\DeclareMathOperator*{\argmax}{arg\;max}

\DeclareMathOperator*{\argtopk}{arg\;top-\mathnormal{k}}

\newcommand{\cmark}{\ding{51}}
\newcommand{\xmark}{\ding{55}}

\newcommand{\trans}{\top}
\newcommand{\hadprod}{\odot}

\newcommand*{\pr}[1]{\left( #1 \right)}
\newcommand*{\br}[1]{\left\{ #1 \right\}}
\newcommand*{\sq}[1]{\left[ #1 \right]}
\newcommand*{\abr}[1]{\left\langle #1 \right\rangle}
\newcommand*{\abs}[1]{\left\vert #1 \right\vert}
\newcommand*{\norm}[1]{\left\Vert #1 \right\Vert}

\newcommand*{\at}[2]{\left. #1 \right|_{#2}}
\newcommand*{\vect}[1]{\mathbf{#1}} %
\newcommand*{\matr}[1]{\mathbf{#1}}

\usepackage{calc}
\newcommand{\coloneq}{:\mathrel{\resizebox{\widthof{$=$}-\widthof{$:$}}{\height}{$=$}}}

\newcommand{\ie}{{i.e.},}
\newcommand{\eg}{{e.g.},}
\newcommand{\etc}{{etc.}}
\newcommand{\registered}{\textsuperscript{\textregistered}}
\newcommand{\tsne}{\mbox{$t$-SNE}}

\newcommand*{\msd}[3][]{{#2}\raisebox{-.2ex}{\small $\pm${#3}}\textsuperscript{#1}}

\newcommand{\mymidline}{\midrule}
\newcommand{\mymidlines}{\midrule\midrule}

\usepackage{amsthm} %
\makeatletter
\NewCommandCopy\latexparagraph\paragraph
\RenewDocumentCommand{\paragraph}{sO{#3}m}{%
    \IfBooleanTF{#1}
    {\latexparagraph*{\maybe@addperiod{#3}}}
    {\latexparagraph[#2]{\maybe@addperiod{#3}}}%
}
\newcommand{\maybe@addperiod}[1]{%
    #1\@addpunct{.}%
}
\makeatother

\DeclareMathAlphabet{\bbold}{U}{bbold}{m}{n}
\newcommand*{\ind}[1]{\bbold{1}\!\sq{#1}} %

\newcommand{\predss}{\left. \hat{y} \middle| x_i \right.}

\newcommand{\documentset}{\mathcal{X}}
\newcommand{\labelset}{\mathcal{Y}}
\newcommand{\lm}{\mathcal{M}_{\vect{\theta}}}
\newcommand{\graph}{\mathcal{G}}
\newcommand{\vertices}{\mathcal{V}}
\newcommand{\edges}{\mathcal{E}}
\newcommand{\selected}{\documentset_s}
\newcommand{\normfac}{\tau}
\newcommand{\clusters}{\mathcal{C}}
\newcommand{\hdbscan}{\textsc{hdbscan*}}
\newcommand{\candsel}{\documentset_c}
\newcommand{\dist}{\Delta}
\newcommand{\Simil}{\Omega}
\newcommand{\simil}{\omega}

\newcommand{\mask}{[$\mathtt{MASK}$]}
\newcommand{\textph}{\textsc{[X]}}
\newcommand{\conph}{\textsc{[domain]}}
\newcommand{\catph}{\textsc{[Y]}}

\newcommand{\mse}{missed cluster effect}

\newcommand{\alps}{\textsc{alps}}

\newcommand{\patron}{\textsc{Patron}}
\newcommand{\kmeans}{$k$-\textsc{means}}

\newcommand{\yelpfull}{Yelp\textsubscript{full}}
\newcommand{\votek}{\textsc{vote}-$k$}

\newcommand{\denoise}{\textsc{Denoise}}%
\newcommand{\knng}{k\mathrm{NN}\!}
\newcommand{\normalize}{\textsc{GraphNorm}}%

\newcommand{\charHeight}{\fontcharht\font`\W}
\newcommand{\inlinepdf}[1]{{\includegraphics[height=\charHeight,width=5\charHeight]{#1}}}

\newcommand{\llama}{\textsc{Llama}~2 7B}

\newcommand{\mymethod}{\mbox{\textsc{Deuce}}}

\title{\mymethod{}: Dual-diversity Enhancement and Uncertainty-awareness\protect{\\}for Cold-start Active Learning}

\author{
  Jiaxin Guo$^{\diamond\dagger\ddagger}$
    \ \ \ %
  C. L. Philip Chen$^{\diamond\dagger\ddagger}$
  \ \ \ %
  Shuzhen Li$^{\dagger\ddagger}$
  \ \ \ %
  Tong Zhang$^{\diamond\dagger\ddagger}$\Thanks{Tong Zhang is the corresponding author.}
  \\
  $^\diamond$Guangdong Provincial Key Laboratory of Computational AI Models and Cognitive Intelligence,\\
  School of Computer Science and Engineering, South China University of Technology, Guangzhou, China
  \\
  $^\dagger$Pazhou Lab, Guangzhou, China
  \\
  $^\ddagger$Engineering Research Center of the Ministry of Education on\\ Health Intelligent Perception and Paralleled Digital-Human, Guangzhou, China
  \\
  \texttt{cs\_guojiaxin@mail.scut.edu.cn}%
  \ \ \ %
  \texttt{philipchen@scut.edu.cn}%
  \\
  \texttt{cslishuzhen@mail.scut.edu.cn}%
  \ \ \ %
  \texttt{tony@scut.edu.cn}%
}

\date{}

\begin{document}
\maketitle
\begin{abstract}
     Cold-start active learning (CSAL) selects valuable instances from an unlabeled dataset for manual annotation.
     It provides high-quality data at a low annotation cost for label-scarce text classification.
     However, existing CSAL methods overlook weak classes and hard representative examples, resulting in biased learning.
     To address these issues, this paper proposes a novel dual-diversity enhancing and uncertainty-aware (\mymethod{}) framework for CSAL.
     Specifically, \mymethod{} leverages a pretrained language model (PLM) to efficiently extract textual representations, class predictions, and predictive uncertainty.
     Then, it constructs a Dual-Neighbor Graph (DNG) to combine information on both textual diversity and class diversity, ensuring a balanced data distribution.
     It further propagates uncertainty information via density-based clustering to select hard representative instances.
     \mymethod{} performs well in selecting class-balanced and hard representative data by dual-diversity and informativeness.
     Experiments on six NLP datasets demonstrate the superiority and efficiency of \mymethod{}.
\end{abstract}

\section{Introduction}
\label{sec:intro}

Cold-start active learning (CSAL; \citealp{yuan-etal-2020-cold,zhang-etal-2022-survey})
has gained much attention for efficiently labeling large corpora from zero. %
Given an unlabeled corpus (\ie{} the ``{cold-start}'' stage), it aims to acquire a small \mbox{subset} (seed set) for annotation. %
Such absence of labels can happen due to \mbox{data privacy} concerns \citep{Holzinger2016,li2023privacy}, limited domain experts\footnote{Recent studies \citep{lu2023human,naeini2023large,zhang2023utilising} have shown that state-of-the-art PLMs still underperform human experts in difficult tasks.} \citep{WU2022364}, labeling difficulty \citep{9650877}, quick expiration of labels \citep{9101545,8654015}, \etc{}
In real-world tasks with specialized domains (\eg{} medical report classification with rare diseases; \citealp{DeAngeli2021}), the complete absence of labels and lack of \textit{a posteriori} knowledge pose challenges to CSAL.

While active learning (AL) has been studied for a wide range of NLP tasks \citep{zhang-etal-2022-survey}, the cold-start problem has been hardly addressed.
At the cold-start stage, the model is untrained and no labeled data are available for validation.
Traditional CSAL applies random sampling \citep{Ash2020Deep,margatina-etal-2021-active},
diversity sampling \citep{8443399,chang-etal-2021-training}, or
uncertainty sampling \citep{schroder-etal-2022-revisiting}.
However, random sampling suffers from high variance \citep{random_instability}; diversity sampling is prone to easy examples and vector space noise \citep{eklund-forsman-2022-topic}; uncertainty sampling is prone to redundant examples, outliers, and unreliable metrics \citep{WOJCIK2022109219}. %
Moreover, existing methods ignore class diversity, where the sampling bias often results in class imbalance \citep{sampling_bias_al}.
At worst, the \emph{\mse{}} \citep{10.1145/1183614.1183709,8443399} can happen, \ie{} clusters of weak classes are neglected. %
\citet{tomanek-etal-2009-proper} showed that an unrepresentative seed set gives rise to this effect.
Learning is misguided, if started unfavorably.

The key challenge for CSAL lies in how to acquire a diverse and informative seed set.
As a general heuristic \citep{DASGUPTA20111767}, a proper seed set should strike a balance between \mbox{{exploring}} the \emph{input space} for instance regions (\eg{} diversity sampling) and {exploiting} the \emph{version space} for decision boundaries (\eg{} uncertainty sampling).
Such hybrid CSAL strategies have been proposed based on combinations of
neighbor-awareness \citep{pmlr-v162-hacohen22a,su2023selective,yu-etal-2023-cold},
\mbox{clustering} \citep{yuan-etal-2020-cold,agarwal2021addressing,10.1007/978-3-031-08473-7_9,brangbour2022cold,shnarch-etal-2022-cluster,yu-etal-2023-cold}, and
uncertainty estimation \citep{dligach-palmer-2011-good,yuan-etal-2020-cold,10.1007/978-3-031-08473-7_9,yu-etal-2023-cold}. %
However, existing methods fail to explore the \emph{label space} to enhance class diversity and mitigate imbalance. %
Moreover, most methods perform diversity sampling followed by uncertainty sampling, treating both aspects in isolation.

\medskip
To address these challenges, this paper presents \mymethod{}, a dual-diversity enhancing and uncertainty-aware framework for CSAL. %
It adopts a graph-based hybrid strategy to enhance diversity and informativeness. %
Different from previous works, \mymethod{} not only emphasizes the diversity in textual contents (textual diversity), but also diversity in class predictions (class diversity).
This is termed \mbox{\emph{\bfseries dual-diversity}} in this paper.
To achieve this in the cold-start stage, it exploits the rich representational and predictive capabilities of PLMs.
For informativeness, the predictive uncertainty is estimated from a one-vs-all (OVA) perspective.
This helps mining informative ``hard examples'' for learning.
Then, \mymethod\ further employs manifold learning techniques \citep{mcinnes2018umap} to derive dual-diversity information.
This results in the novel construction of a Dual-Neighbor Graph (DNG).
Finally, \mymethod\ performs density-based uncertainty propagation and Farthest Point Sampling (FPS) on the DNG.
While propagation prioritizes \emph{\bfseries representatively uncertain} (RU) instances, FPS enhances the dual-diversity.
Overall, \mymethod{} ensures a more diverse and informative acquisition.

The merits of \mymethod{} are attributed to the following contributions:
\begin{itemize}[nosep]
    \item
    The dual-diversity enhancing and uncertainty aware (\mymethod{}) framework adopts a novel hybrid acquisition strategy.
    It effectively selects class-balanced and hard representative instances, achieving a good balance between exploration and exploitation in CSAL.

    \item
    This paper proposes a graph-based dual-diversity enhancement mechanism to select diverse instances with textual diversity and class diversity, tackling class imbalance in CSAL.

    \item
    This paper presents an embedding-based uncertainty-aware prediction mechanism to effectively select hard representative instances according to predictive uncertainty.

\end{itemize}

\section{Related Work}
\label{sec:related-work}

\subsection{Cold-start Active Learning (CSAL)}
\label{sec:csal-taxonomy}

According to the taxonomy of \citet{zhang-etal-2022-survey}, CSAL research for NLP can be categorized as informativeness-based, representativeness-based, and hybrid.
As most methods are hybrid, the techniques and challenges for informativeness or representativeness are elucidated below.

\subsubsection{Informativeness}
\label{subsubsec:informativeness}

\paragraph{Uncertainty}
The main metric for informativeness in CSAL is uncertainty, as it is more tractable in cold-start stages than others (\eg{} gradients).
High predictive uncertainty indicates difficulty for the model, thus valuable for annotation.
Most existing methods use language models (LMs) for estimation.
Common estimators include
entropy \citep{zhu-etal-2008-active,yu-etal-2023-cold},
LM probability \citep{dligach-palmer-2011-good},
LM loss \citep{yuan-etal-2020-cold}, and
probability margin \citep{10.1007/978-3-031-08473-7_9}.
However, several challenges exist in uncertainty estimation:
(a)
Often, a closed-world assumption is imposed.
In other words, predictions are normalized such that they sum to $1$.
This hinders the expression of uncertainty, as it forces mapping to one of the known classes, ignoring options such as ``none of the above'' \citep{ova-unc}.
(b) PLMs suffer from overconfidence \citep{park-caragea-2022-calibration,wang2023calibration}.
This requires calibration for more robust uncertainty estimation \citep{yu-etal-2023-cold}.
(c)
Task information is hardly considered.
As a result, the uncertainty will not be related to the downstream task (output uncertainty), but rather its intrinsic perplexity (input uncertainty) \citep{jiang-etal-2021-know}.
\patron{} \citep{yu-etal-2023-cold} uses task-related prompts to tackle this issue.

\subsubsection{Representativeness}
\label{subsubsec:repr}

\paragraph{Density}
To avoid outliers, density-based CSAL methods prefer ``typical'' instances.
The method of \citet{zhu-etal-2008-active} and TypiClust (\citealp{pmlr-v162-hacohen22a}) prioritize instances with high $k$NN density.
Uncertainty propagation \citep{yu-etal-2023-cold} is also useful in aggregating density information.
A typical group of uncertain examples indicates a region where the model's knowledge is lacking.

\paragraph{Discriminative}
Some CSAL methods acquire sequentially or iteratively.
They thus discriminate, \ie{} prefer an instance if it differs the most from selected ones.
Coreset selection \citep{sener2018active} selects an instance (cover-point) such that its minimum distance to selected instances is maximized.
\votek{} \citep{su2023selective} adopts a greedy approach to select remote instances on a $k$NN graph.

\paragraph{Batch diversity}
It is more efficient to acquire in batch mode \citep{settles.tr09}, \ie{} to select multiple instances at each step.
Clustering has been a common technique to enhance batch diversity and avoid redundancy in CSAL.
It helps structure the unlabeled dataset by grouping similar instances together.
\citet{10.1145/1015330.1015349} and \citet{10.1007/978-3-540-24775-3_46} first proposed \mbox{pre-clustering} the input space to select representatives from each cluster. %
\citet{dasgupta-ng-2009-mine} used spectral clustering on the similarity matrix of documents.
\citet{hu2010off} and \citet{8443399} used hierarchical clustering to stabilize the process.
\citet{zhu-etal-2008-active} and more recent works \citep{yuan-etal-2020-cold,chang-etal-2021-training,agarwal2021addressing,10.1007/978-3-031-08473-7_9,pmlr-v162-hacohen22a,yu-etal-2023-cold} have commonly used \kmeans{} for its simplicity and efficiency.
However, these clustering methods can be sensitive to outliers.
Moreover, clustering in the input space only contributes to textual diversity, regardless of other aspects.%

\subsection{Missed Cluster Effect}
The \mse{} \citep{10.1145/1183614.1183709,tomanek-etal-2009-proper} is an extreme case of class imbalance.
It refers to when an AL strategy neglects certain classes (or clusters within classes).
\citet{10.1145/1183614.1183709} first recognized the \mse{} in the context of text classification.
They suggested more use of domain knowledge. %
Knowledge extraction from PLMs is in harmony with this suggestion.
\citet{dligach-palmer-2011-good} proposed an uncertainty-based approach to avoid the \mse{} in word sense disambiguation (WSD). %
However, it is based on task-agnostic LM probability.
\citet{marcheggiani-artieres-2014-experimental} showed that labeling relevant instances, which reduces the labeling noise, also helps mitigate the \mse{}.
Label calibration aligns with this finding.
While many works are devoted to addressing the \mse{} or general class imbalance (\eg{} \citealp{9093475,fairstein-etal-2024-class}) for general AL, they often rely on a labeled subset. %
Class diversity enhancement would help mitigate class imbalance issues, but it remains an open question for CSAL.

\section{Methodology}
\label{sec:csac}

In this section, the methodology of the proposed \mymethod{} is introduced.
Section~\ref{subsec:problem-formulat} first defines CSAL and declares the notations for the rest of this paper.
The framework of \mymethod{} is then elaborated in Section~\ref{subsec:overview-method}.

\subsection{Problem Formulation}
\label{subsec:problem-formulat}

This paper considers CSAL in a pool-based manner. %
Learning is initiated with a set of $N$ unlabeled documents, $\documentset \coloneq \br{x_i}_{i=1}^N$.
A $C$-way text classification task is defined by a set of classes $\labelset \coloneq \br{y_j}_{j=1}^C$ taking values in a domain %
 $\mathbb{Y}$.

Given a labeling budget $b \ll N$, a CSAL strategy acquires a subset $\selected \subset \documentset$ with a fixed size $\abs{\selected} = b$, such that the labeled subset $\selected'$ boosts most performance when used as a training seed set.
The performance is evaluated by fine-tuning a PLM $\lm$ with $\selected'$, and testing for its accuracy.

\begin{figure*}[tbh]
    \includegraphics[width=\textwidth]{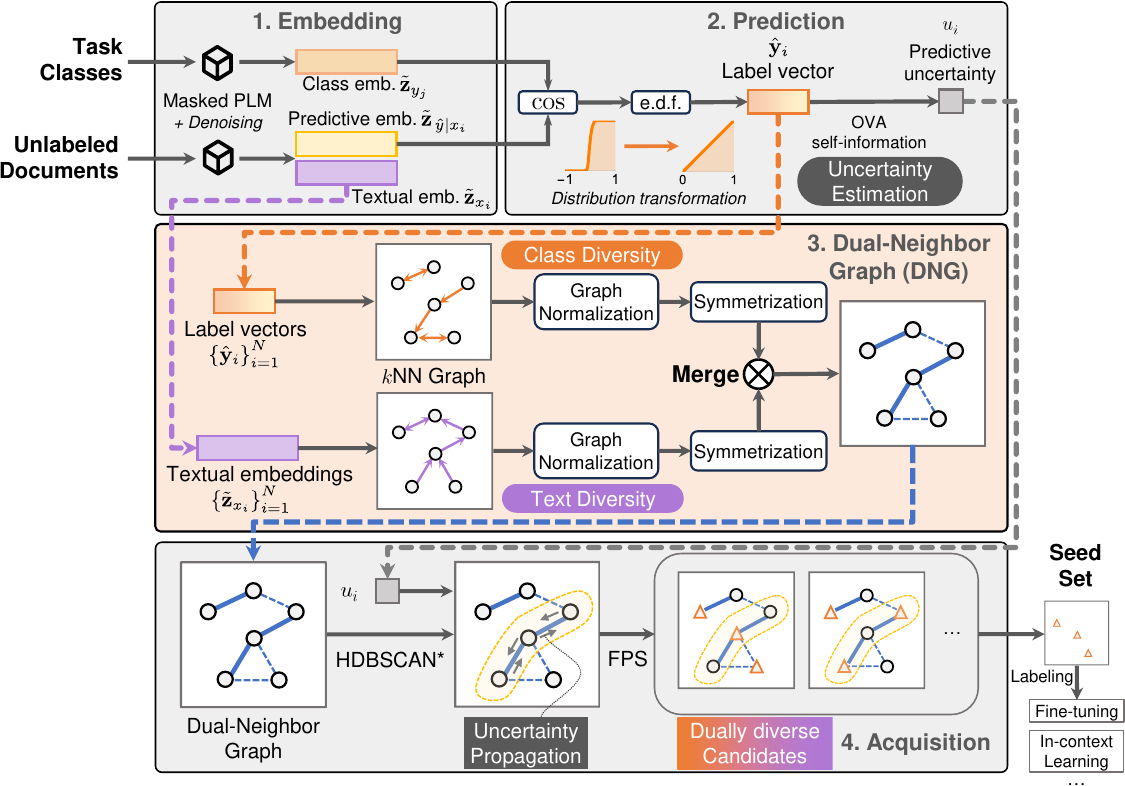}
    \caption{The proposed \mymethod{} framework.}
    \label{fig:diagram}
\end{figure*}

\subsection{The \mymethod{} Framework}
\label{subsec:overview-method}

The proposed \mymethod{} framework is illustrated in Figure~\ref{fig:diagram}.
Overall, the components of \mymethod\ serve the same goal---to produce a seed set with high dual-diversity and informativeness.

\subsubsection{Embedding Module}
\label{subsec:prompt-pred-unc}

In CSAL, data selection starts with only an \mbox{unlabeled} corpus.
\mymethod\ leverages PLM embeddings, which guide the selection process towards more diverse and informative samples.

Specifically, the embedding module implements a {prompt-based, verbalizer-free} approach \citep{jiang-etal-2022-promptbert}.
This requires only a single inference pass per document.%

\paragraph{Textual and predictive embedding}
In a masked PLM, the bidirectional semantics can be condensed into a \mask{} token.
In light of this, \mymethod{} extends \citet{jiang-etal-2022-promptbert}'s template with double \mask{} tokens:
$$
T_x \coloneq \fbox{%
\begin{varwidth}{\linewidth}%
    This sentence: ``\textph{}'' means \mask{}.\\
    Its \conph{} is \mask{}.
\end{varwidth}%
} \text{,}
$$
where \conph{} is the target domain $\mathbb{Y}$, such as ``sentiment''.
The hidden representations of \mask{} tokens are extracted as the textual $\vect{z}_{x_i}$ and predictive embeddings $\vect{z}_{\predss}$. %
They capture the intrinsic and task-related semantics.

However, raw embeddings suffer from template bias and length bias \citep{10.1145/3583780.3614833}.
\mymethod{} further applies \emph{template denoising} (\citealp{jiang-etal-2022-promptbert}%
) to obtain the denoised embeddings $\tilde{\vect{z}}$.

\paragraph{Class embedding}
Predictions need to be paired with the known classes.
Class embeddings $\tilde{\vect{z}}_{y_j}$ are generated from a prompt template $T_y$, similar to $T_x$:
$$
T_y \coloneq \fbox{This \conph{}: ``\catph{}'' means \mask{}.} \text{,}
$$
where \catph{} is the placeholder for a class $y_j$.

\subsubsection{Prediction Module}
\label{subsec:predict-module}

This module aims to produce uncertainty-aware labels.
With class information, \mymethod{} gains prior knowledge about potential data distributions.
With uncertainty information, \mymethod{} is informed of potential labeling gain.

\paragraph{Label vector}
For better uncertainty estimation,
\mymethod{} adopts a One-vs-All (OVA) setup, such that labels $\hat{\vect{y}}_i$ do not necessarily sum to $1$.
First, it computes the \mbox{inner product} $\simil_{ij}$ for each pair of predictive and class embeddings:
\begin{align*}
    \matr{\Simil} & = \begin{bmatrix}
        \tilde{\vect{z}}_{\hat{y} | \mysub{x}{1}} &
        \cdots &
        \tilde{\vect{z}}_{\hat{y} | \mysub{x}{N}}
    \end{bmatrix}^\trans \begin{bmatrix}
        \tilde{\vect{z}}_{\mysub{y}{1}} &
        \cdots &
        \tilde{\vect{z}}_{\mysub{y}{C}}
    \end{bmatrix} \\
    & \coloneq \begin{bmatrix}
        \simil_{ij}
    \end{bmatrix}_{i=1,j=1}^{N,C} \text{.}
\end{align*}

Ideally, similarity $\simil_{ij}$ can be linearly transformed to class label $\hat{y}_{ij}$.
However, high anisotropy \citep{gao2018representation} was observed in preliminary experiments. %
As a result, $\simil_{ij}$ has a non-uniform distribution over $[-1,1]$.
To tackle this issue, %
\mymethod{} uses the empirical distribution function (e.d.f.) of $\matr{\Simil}$ to give a calibrated estimate of labels $\hat{\matr{Y}}$:%
\begin{align*}
    \hat{y}_{ij} = \hat{\mathbb{F}}_{\matr{\Simil}}\!\pr{\simil_{ij}} = \frac{1}{NC} \sum_{m=1}^N \sum_{n=1}^C \ind{\simil_{mn} \le \simil_{ij}} \text{,}
\end{align*}
where $\ind{\cdot}$ is the indicator function.
This gives $\hat{y}_{ij} \sim U(0,1)$ regardless of the embedding distribution.

\paragraph{Predictive uncertainty}
In CSAL, uncertainty represents the difficulty of an instance.
\mymethod\ adapts entropy, a common measure of uncertainty (\S\ref{subsubsec:informativeness}).

In information theory, entropy is the expected self-information $I$ of possible events.
In an OVA setup, possible events $\br{E_{i}}$ are ``$x_i$ has a high predictive score for \emph{exactly one} class''.
The probability of event $E_i$ is given by \citet{WOJCIK2022109219}:
\begin{align*}
    p(E_i) = \max_{j} { \hat{y}_{ij} \prod_{\substack{l=1 \\ l\ne j}}^C \pr{1-\hat{y}_{il}} } \text{.}
\end{align*}
Therefore, \mymethod{} adopts the entropy from $\br{E_i}$ as the uncertainty estimate $\vect{u}$:
\begin{align*}
    u_i = I(E_i) = - \log p(E_i) \text{.}
\end{align*}

\subsubsection{Dual-Neighbor Graph (DNG) Module}
\label{subsec:dng}

Graphs serve as a powerful tool for data selection by explicitly modeling data interrelationship.
This enables the propagation of valuable information (\eg{} uncertainty) and the selection of more diverse samples.
To integrate textual and class diversity, \mymethod{} leverages manifold learning techniques \citep{mcinnes2018umap} on $k$-Nearest-Neighbor ($k$NN) graphs of both spaces.\footnote{It is worth noting that \mymethod{} does not utilize or optimize any Graph Neural Network (GNN). With the rich representational capability of PLMs, \mymethod{} does not require GNNs to learn data representations.}

\paragraph{$k$NN graph}

The use of $k$NN arises from the neighborhood perspective of diversity.
\mymethod{} aims to avoid selecting neighboring instances.
In a $k$NN graph, an instance $x_i$ is connected with its $k$ nearest neighbors $\br{\mysub{x_i}{j}}$ under some \mbox{distance function $\dist(\cdot, \cdot)$}.
Formally, the two metric spaces of $k$NN are defined as follows.
\begin{itemize}[noitemsep]
    \item The textual space $(\documentset, \dist_{\tilde{z}})$ is defined by textual embeddings under cosine distance, $\dist_{\tilde{z}}(x_i,x_j) = \smash[t]{ \frac{1}{\pi} \arccos\!\pr{\tilde{\vect{z}}_{x_i}^{\trans} \tilde{\vect{z}}_{x_j}^{\vphantom{\trans}}} }$;
    \item The label space $(\documentset, \dist_{\hat{y}})$ is defined by label vectors under $\ell_1$ distance, $\dist_{\hat{y}} (x_i,x_j) = \norm{\hat{\vect{y}}_i - \hat{\vect{y}}_j}_1$.
\end{itemize}
The $k$NN graph from each space is denoted by $\graph_{\tilde z}$ and $\graph_{\hat y}$, respectively.

\paragraph{Graph normalization}
To unify textual and class diversity, \mymethod{} merges the two $k$NN graphs into one for graph-based sampling.
However, across two distinct spaces, it is necessary to first normalize the distances \citep{mcinnes2018umap}.

To ease notation, this part omits the subscript as $\graph \in \br{\graph_{\tilde z}, \graph_{\hat y}}$.
For each $x_i$, \mymethod{} finds a normalization factor $\normfac_i > 0$ that satisfies the equation
\begin{align*}
    \sum_{j=1}^k \exp\!\pr{-\frac{\dist\!\pr{x_i,\mysub{x_i}{j}} - \rho_i}{\normfac_i}} = \log_2\!k \text{,}
\end{align*}
where $\rho_i$ denotes $x_i$'s distance to its nearest neighbor.
The weights $\tilde{w}$ of the normalized (directed) $k$NN graph $\graph$, denoted by $\tilde{\graph}$, is defined by
\begin{align*}
    \tilde{w}\!\pr{\abr{x_i,\mysub{x_i}{j}}} \coloneq \exp\!\pr{-\frac{\dist\!\pr{x_i,{x_i}_{{}_j}} - \rho_i}{\normfac_i}} . %
\end{align*}
After normalization, the original $k$NN weights $w \in [0, \infty)$ are transformed to $\tilde{w} \in (0,1]$.

\paragraph{Symmetrization}
To identify representative instances, \mymethod{} performs graph clustering.
This requires symmetric $k$NN graphs.

Let $\tilde{\matr{W}}$ denote the sparse weight matrix of $\tilde{\graph}$.
Since weights $\tilde{w} \in [0,1]$, they can be interpreted as fuzzy memberships of neighborhood.
Hence, symmetrizing $\tilde{\matr{W}}$ is equivalent to finding the fuzzy union \citep{dubois:hal-04067331} of the neighbors $\tilde{\matr{W}}$ and reverse neighbors $\tilde{\matr{W}}^\trans$:
\begin{align*}
\tilde{\matr{W}}_\text{sym} = \tilde{\matr{W}} + \tilde{\matr{W}}^\trans - \tilde{\matr{W}} \hadprod \tilde{\matr{W}}^\trans \text{,}
\end{align*}
where $\hadprod$ is the Hadamard product.
$\tilde{\matr{W}}_\text{sym}$ defines the weights of the symmetric $k$NN graph $\tilde{\graph}_\text{sym}$.
Its edges are denoted by $\tilde{\edges}_\text{sym}$.

\newcommand{\myedge}{\br{x_i,x_j}}
\newcommand{\ezsym}{\tilde{\edges}_{\tilde{z}, \text{sym}}}
\newcommand{\eysym}{\tilde{\edges}_{\hat{y}, \text{sym}}}
\newcommand{\wzsym}{\tilde{w}_{\tilde{z}, \text{sym}}}
\newcommand{\wysym}{\tilde{w}_{\hat{y}, \text{sym}}}
\newcommand{\myindent}{\hphantom{\wzsym}}
\paragraph{Merging}
It is now appropriate to merge the two $k$NN graphs.
This unifies textual and class diversity in one graph.

As merged, the {Dual-Neighbor Graph} (DNG) is an undirected graph $\graph_\text{dual} = (\vertices, \edges, w_\text{dual})$.
The edges $\edges$ are the union of edges in $\tilde{\graph}_{\tilde{z}, \text{sym}}$ and $\tilde{\graph}_{\hat{y}, \text{sym}}$.
Moreover, $\edges$ is divided into two types:
\begin{itemize}[nosep]
    \item $\edges_1$ represents edges which only appear in \mbox{either} $k$NN graph, called \emph{single-neighbor edges};
    \item $\edges_2$ represents edges which appear in both $k$NN graphs, called \emph{dual-neighbor edges}. They connect neighboring documents which are similar in {both} textual semantics and class predictions.
\end{itemize}
The weight $w_\text{dual}$ of an undirected edge $\myedge \in \mathcal{E}$ is thereby defined as
\begin{align*}
    w_\text{dual} \coloneq \begin{cases}
        \wzsym \wysym + \gamma \\ \myindent \text{if }
        \myedge \in \edges_2 \text{,} \\
        \wzsym \\ { \myindent \text{if }
        \myedge \in \smash[b]{\color{gray}\underbrace{\color{black} \ezsym \setminus \eysym }_{ \subset~\edges_1 } }} \text{,} \\
        \wysym  \\ \myindent \text{if }
        \myedge \in \smash[b]{\color{gray}\underbrace{\color{black} \eysym \setminus \ezsym }_{ \subset~\edges_1 }} \text{;}
    \end{cases}
    \vphantom{%
    \begin{cases}%
        \wzsym \wysym + \gamma \\ \myindent \text{if }
        \myedge \in \edges_2 \\
        \wzsym \\ { \myindent \text{if }
        \myedge \in \ezsym \setminus \eysym \subset \edges_1} \\
        \wysym  \\ \myindent \text{if }
        \myedge \in \underbrace{ \eysym \setminus \ezsym }_{ \subset \edges_1 }
    \end{cases}
    }
\end{align*}
where $\gamma$ is a threshold to distinguish dual-neighbor edges $\edges_2$ from single-neighbor edges $\edges_1$.
In essence, DNG assigns greater weights to dual-neighbor edges.
As a result, during the subsequent graph clustering and traversal, \mymethod{} can avoid selecting textual and class neighbors.

\subsubsection{Acquisition Module}
\label{subsec:dsdc}

\mymethod{} adopts a hybrid acquisition strategy.
Overall, the goal is to produce a diverse and informative seed set.
To achieve this, the acquisition module performs graph clustering, propagation, and traversal on DNG.

\paragraph{\hdbscan{}}

A group of similar documents with high predictive uncertainty indicates an area where the model's knowledge is lacking.
By labeling one of the documents, the model predictions can be improved for similar ones in the area.
Therefore, it is valuable to identify and prioritize such {representatively uncertain} (RU) groups for CSAL.

Clustering has been a common technique to group similar instances (\S\ref{subsubsec:repr}).
However, traditional clustering methods (\eg\ \kmeans{}) are ill-suited, as the number of RU groups is unknown.
Moreover, they force every instance into a cluster, while some instances may not belong to any RU group.
Instead, \mymethod{} adopts density-based clustering, %
which identifies RU groups with a sufficient density ($\ge k_r$ similar documents).

Specifically, \mymethod{} applies \hdbscan{} \citep{10.1007/978-3-642-37456-2_14,10.1145/2733381} on the DNG, with minimum cluster size $k_r$. %
A document $x_i$ is either (a) clustered in an RU group $c_l$ with membership $p_i$, or (b) excluded as a non-RU outlier.

\paragraph{Uncertainty propagation}
To prioritize RU documents, uncertainty information (\S\ref{subsec:predict-module}) is propagated and aggregated in RU groups.
This is formulated as a single step of message propagation:
\begin{align*}
    \tilde{u}_i = u_i + \sum_{x_j \in c_l \setminus \br{x_i}}{w_{\text{dual}}\!\pr{\br{x_i,x_j}} p_j u_j} \text{.}
\end{align*}

\paragraph{FPS}
The final acquisition adopts a combination of diversity sampling and uncertainty sampling.
First, \mymethod{} runs Farthest Point Sampling (FPS; \citealp{577129}) on the DNG. %
As the result only depends on the initial point, FPS is started from documents $x_i$ with top-$k$ degrees.
Each produces a candidate seed set $\smash[t]{\candsel^{(i)}}$, which contains $b$ dually diverse samples.
Finally, \mymethod{} chooses the candidate with the highest propagated uncertainty:
\begin{align*}
    \selected = \argmax_{\candsel^{(i)}} \sum_{x_j \in \candsel^{(i)}} \tilde{u}_j.
\end{align*}

The whole process is described in Algorithm~\ref{alg:cs}.
\begin{algorithm}[h!]
    \caption{Cold-start acquisition in \mymethod{}.}
    \label{alg:cs}
    \begin{algorithmic}[1]
        \Require{\small unlabeled documents $\documentset$, classes $\labelset$, labeling budget $b$, number of neighbors $k$, representativeness threshold $k_r$, and frozen PLM $\lm$.}
        \LComment{\footnotesize Embedding (\S\ref{subsec:prompt-pred-unc}) and prediction (\S\ref{subsec:predict-module}).}
        \ForAll{$y_j \in \labelset$}
            \State{$\tilde{\vect{z}}_{y_j} \gets \denoise\!\pr{ \lm\!\pr{\at{T_y}{y_j}} }$}
        \EndFor
        \ForAll{$x_i \in \documentset$}
            \State{$\tilde{\vect{z}}_{x_i}, \tilde{\vect{z}}_{\predss} \gets \denoise\!\pr{ \lm\!\pr{\at{T_x}{x_i,\mathbb{Y}}} }$}
            \ForAll{$y_j \in \labelset$}
                \State{$\simil_{ij} \gets \tilde{\vect{z}}_{\predss}^\trans \tilde{\vect{z}}_{y_j}^{\vphantom{\trans}}$}
                \State{$\hat{y}_{ij} \gets \hat{\mathbb{F}}_{\matr{\Simil}}\!\pr{\simil_{ij}}$}%
            \EndFor
            \State{$u_i \gets -\log p(E_i')$} \Comment{\footnotesize Uncertainty estimation.}%
        \EndFor
        \LComment{\footnotesize Dual-Neighbor Graph (\S\ref{subsec:dng}).}
        \State{$\tilde{\graph}_{\tilde{z},\text{sym}} \gets \normalize\!\pr{\knng\pr{ {
        \documentset , \dist_{\tilde{z}} } }}$}
        \State{$\tilde{\graph}_{\hat{y},\text{sym}} \gets \normalize\!\pr{\knng\pr{ {
        \documentset , \dist_{\hat{y}} } }}$}
        \State{$\graph_\text{dual} \gets \mathrm{DNG}\!\pr{ \tilde{\graph}_{\tilde{z},\text{sym}} , \tilde{\graph}_{\hat{y},\text{sym}}; \gamma }$}
        \LComment{\footnotesize  Acquisition (\S\ref{subsec:dsdc}).}
        \State{$\clusters \gets \text{\hdbscan}\!\pr{\graph_\text{dual}; k_r}$}
        \ForAll{$x_i \in \documentset$}
            \If{$\exists c_l \in \clusters : x_i \in c_l$}
                \State{$\tilde{u}_i \gets \textsc{Propagate}\!\pr{u_i, c_l}$}%
            \EndIf
        \EndFor
        \ForAll{$x_i \in \argtopk{\mathsf{deg}(x_i, \graph_\text{dual})} $}
            \State{$\candsel^{(i)} \gets \mathrm{FPS}\!\pr{\graph_\text{dual},x_i;b}$}
        \EndFor
        \Return{$\selected \gets \argmax_{\candsel^{(i)}} \sum_{x_j \in \candsel^{(i)}} \tilde{u}_j$}
        \Ensure{\small A dually diverse and informative seed set $\selected \subset \documentset$.}
    \end{algorithmic}
\end{algorithm}

\begin{table*}[t!]
\centering
    \begin{adjustbox}{width=\textwidth}
        \small
\begin{tabular}{lllcccl}
\toprule
\textbf{Dataset} & \textbf{Source domain} & \textbf{Target domain~$\mathbb{Y}$} & \textbf{\#Class~$C$} & \textbf{\#Unlabeled~$\abs{\documentset}$} & \textbf{\#Test} & \textbf{Label distribution (bar chart) and names~$y_j$} \\
\midrule
\vcell{IMDb} & \vcell{Movie review} & \vcell{Sentiment} & \vcell{2} & \vcell{25,000} & \vcell{25,000} & \vcell{\inlinepdf{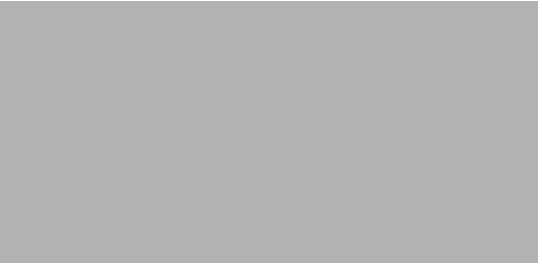} Negative, Positive} \\[-\rowheight]
\printcelltop & \printcelltop & \printcelltop & \printcelltop & \printcelltop & \printcelltop & \printcelltop \\
\vcell{\yelpfull} & \vcell{Review} & \vcell{Rating} & \vcell{5} & \vcell{38,352} & \vcell{10,000} & \vcell{\inlinepdf{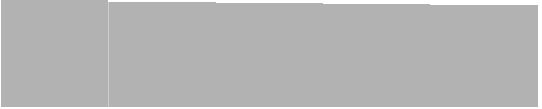} 1 star, 2 stars, 3 stars, 4 stars, 5 stars} \\[-\rowheight]
\printcelltop & \printcelltop & \printcelltop & \printcelltop & \printcelltop & \printcelltop & \printcelltop \\
\vcell{AG's News} & \vcell{News} & \vcell{Category} & \vcell{4} & \vcell{120,000} & \vcell{7,600} & \vcell{\inlinepdf{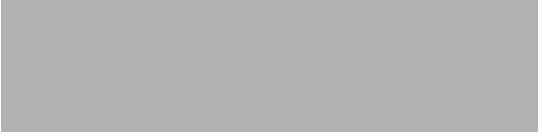} World, Sports, Business, Sci/Tech} \\[-\rowheight]
\printcelltop & \printcelltop & \printcelltop & \printcelltop & \printcelltop & \printcelltop & \printcelltop \\
\vcell{Yahoo! Answers} & \vcell{Web Q\&A} & \vcell{Category} & \vcell{10} & \vcell{300,000$^\dagger$} & \vcell{60,000} & \vcell{\begin{tabular}[b]{@{}l@{}}\inlinepdf{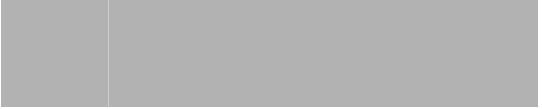} Society \& Culture, Science \& Mathematics, Health,\\\quad{}Education \& Reference, Computers \& Internet, Sports,\\\quad{}Business \& Finance, Entertainment \& Music,\\\quad{}Family \& Relationships, Politics \& Government\end{tabular}} \\[-\rowheight]
\printcelltop & \printcelltop & \printcelltop & \printcelltop & \printcelltop & \printcelltop & \printcelltop \\
\vcell{DBpedia} & \vcell{\begin{tabular}[c]{@{}l@{}}Wikipedia lead\\section\end{tabular}} & \vcell{Category} & \vcell{14} & \vcell{420,000$^\dagger$} & \vcell{70,000} & \vcell{\begin{tabular}[b]{@{}l@{}}\inlinepdf{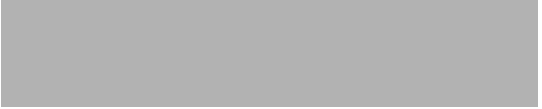} Company, Educational institution, Artist, Athlete,\\\quad{}Office holder, Mean of transportation, Building,\\\quad{}Natural place, Village, Animal, Plant,\\\quad{}Album, Film, Written work\end{tabular}} \\[-\rowheight]
\printcelltop & \printcelltop & \printcelltop & \printcelltop & \printcelltop & \printcelltop & \printcelltop \\
\vcell{TREC} & \vcell{Question} & \vcell{Category} & \vcell{6} & \vcell{5,452} & \vcell{500} & \vcell{\begin{tabular}[b]{@{}l@{}}\inlinepdf{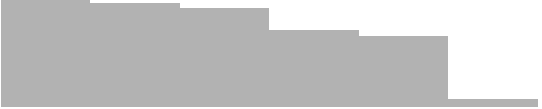}$^\ddagger$ Abbreviation, Entity, Description and abstract concept,\\\quad{}Human being, Location, Numeric value\end{tabular}} \\[-\rowheight]
\printcelltop & \printcelltop & \printcelltop & \printcelltop & \printcelltop & \printcelltop & \printcelltop \\
\bottomrule
\end{tabular}
\end{adjustbox}
    \caption{
        Statistics of evaluation datasets.
        $^\dagger$\emph{Yahoo!} and \emph{DBpedia} are the truncated version with 30k samples per class by \citet{yu-etal-2023-cold}.
        $^\ddagger$\textit{TREC} is an imbalanced dataset.
    }
    \label{table:dataset-stats}
\end{table*}

\section{Experiments and Results}
\label{sec:exp}

\subsection{Experimental Setup}

\paragraph{Datasets}
\label{subsec:datasets}

\mymethod{} is evaluated on six text classification datasets:
{IMDb} \citep{maas-etal-2011-learning},
{\yelpfull} \citep{Meng_Shen_Zhang_Han_2019},
{AG's News} \mbox{\citep{NIPS2015_250cf8b5}},
{{Yahoo}! Answers} \citep{NIPS2015_250cf8b5},
{DBpedia} \citep{Lehmann2015}, and
{TREC} \citep{li-roth-2002-learning}.
Dataset statistics are shown in Table~\ref{table:dataset-stats}.
All the datasets used in the experiments are publicly accessible.
The original labels are removed to create a cold-start scenario.

\paragraph{Evaluation metric}

To evaluate the performance of the acquired seed set $\selected$, it is labeled and used for fine-tuning the PLM.
The original labels of the seed set are revealed.
The accuracy of the fine-tuned PLM on the test set is then reported.
To be consistent with previous methods \citep{yu-etal-2023-cold}, the experiments adopt RoBERTa-base \citep{liu2019roberta} as the backbone PLM. %

\paragraph{Analysis metrics}

To analyze the effect of dual-diversity enhancement, the class imbalance (IMB) and textual-diversity value of seed sets are reported.
Both metrics are computed under budget $b=128$.
IMB \citep{yu-etal-2023-cold} is defined as:
\begin{align*}
    \mathsf{IMB} = \frac{\max_{j=1}^C n_j}{\min_{j=1}^C n_j} \text{,}
\end{align*}
where $n_j$ is the number of instances from class $y_j$.
Textual-diversity value \citep{ein-dor-etal-2020-active,yu-etal-2023-cold} is defined as:
\begin{align*}%
    D = \pr{\frac{1}{ \abs{\documentset \setminus \selected} } \sum_{x_i \in \documentset \setminus \selected}{ \min_{x_j \in \selected} {\Delta\!\pr{x_i,x_j}} }}^{-1} \text{,}
\end{align*}
where $\Delta\!\pr{x_i,x_j}$ is the Euclidean distance of SimCSE embeddings \citep{gao-etal-2021-simcse} of $x_i$ and $x_j$.

\paragraph{Implementation details}
The fine-tuning setup and hyperparameters are the same as \patron{} \citep{yu-etal-2023-cold}'s.
Notably, the experiment code transplants the original implementation of graph normalization (\citealp{McInnes2018}) to GPU for acceleration.
For \mymethod{}, $k=500$, $k_r=3$, and $\gamma=1.0$ (since $\tilde{w}_\text{sym} \le 1.0$) are taken.
All experiments are run on a machine with a single \textsc{nvidia}\registered{} A800 GPU with 80~GB of VRAM.

\paragraph{Baselines}
\label{subsec:baselines}

The following CSAL baseline methods are considered:
\begin{table}[t]
\centering
    \begin{adjustbox}{width=\columnwidth}
        \scriptsize
\begin{tabular}{rcccc}
    \toprule
    \multicolumn{1}{c}{\multirow{2}{*}{Method}} & Informativeness & \multicolumn{3}{c}{Representativeness} \\
    \cmidrule(lr){2-2}
    \cmidrule(lr){3-5}
    \multicolumn{1}{c}{} & Uncertainty & Density & \begin{tabular}[c]{@{}c@{}}Textual\\diversity\end{tabular} & \begin{tabular}[c]{@{}c@{}}Class\\diversity\end{tabular} \\
    \midrule
    Random & \xmark{} & \xmark{} & \xmark{} & \xmark{} \\
    Entropy & \cmark{} & \xmark{} & \xmark{} & \xmark{} \\
    Coreset & \xmark{} & \xmark{} & \cmark{} & \xmark{} \\
    \alps{} & \cmark{} & \xmark{} & \cmark{} & \xmark{} \\
    \textsc{few-s.} & \xmark{} & \xmark{} & \cmark{} & \xmark{} \\
    TypiCl. & \xmark{} & \cmark{} & \cmark{} & \xmark{} \\
    \patron & \cmark{} & \cmark{} & \cmark{} & \xmark{} \\
    \votek & \xmark{} & \cmark{} & \cmark{} & \xmark{} \\
    \midrule
    \mymethod{} & \cmark{} & \cmark{} & \cmark{} & \cmark{} \\
    \bottomrule
\end{tabular}
\end{adjustbox}
    \caption{Comparisons of CSAL methods, which adapt the taxonomy of \citet{zhang-etal-2022-survey} (\S\ref{sec:csal-taxonomy}).}
    \label{table:baselines}
\end{table}

\begin{itemize}[nosep]
    \item \textbf{Random} sampling selects uniformly.%
    \item \textbf{Entropy}-based {uncertainty} sampling (revisited by \citealp{schroder-etal-2022-revisiting})
    selects data with the highest predictive entropy.
    \item \textbf{Coreset} selection \citep{sener2018active}
    iteratively selects data whose minimum distance to the selected data is maximized.
    \item \textbf{\alps} \citep{yuan-etal-2020-cold}
    computes \emph{surprisal embeddings} from BERT loss as uncertainty. They are then clustered with \kmeans{}. %
    Data closest to each centroid are selected.
    \item \textbf{\textsc{few-selector}} (\citealp{chang-etal-2021-training})
    clusters the text embeddings with \kmeans{}.%
    \item \textbf{TypiClust} (\citealp{pmlr-v162-hacohen22a})
    clusters the text embeddings with \kmeans, and selects data with the highest {typicality}, \ie{} $k$NN density, from each cluster.
    \item \textbf{\patron} \citep{yu-etal-2023-cold}
    clusters the text embeddings with \kmeans{}, and selects from each cluster data with the highest propagated uncertainty.
    It then iteratively updates the set to refine inter-sample distances.
    \item \textbf{\votek} \citep{su2023selective}
    iteratively assigns a high score if a data is far from selected data.%
\end{itemize}
Comparisons of the CSAL baselines and \mymethod{} are presented in Table~\ref{table:baselines}.

\subsection{Accuracy Improvement}
\begin{table*}[t!]
    \centering
    \begin{adjustbox}{width=\textwidth}
\begin{tabular}{cc|ccc|ccccc|c}
    \toprule
    \textbf{Dataset} & $b$ & \textbf{Random} & \textbf{Entropy} & \textbf{Coreset} & \textbf{\alps} & \textbf{\textsc{few-s.}} & \textbf{TypiCl.} & \textbf{\patron} & \textbf{\votek} & \textbf{\textsc{\mymethod}} \\
    \mymidlines
    \multirow{3}{*}{IMDb} & 32 & \msd{80.2}{2.5} & \msd{81.9}{2.7} & \msd{74.5}{2.9} & \msd{82.2}{3.0} & \msd{79.2}{1.6} & \msd{82.8}{2.2} & \msd{85.5}{1.5} & \msd{\underline{85.6}}{1.8} & \msd{\bf 86.9}{\bf 0.9} \\
    & 64 & \msd{82.6}{1.4} & \msd{84.7}{1.5} & \msd{82.8}{2.5} & \msd{86.1}{0.9} & \msd{84.9}{1.5} & \msd{84.0}{0.9} & \msd{87.3}{1.0} & \msd{\underline{88.0}}{1.2} & \msd{\bf 88.5}{\bf 0.7} \\
    & 128 & \msd{86.6}{1.7} & \msd{87.1}{0.7} & \msd{87.8}{0.8} & \msd{87.5}{0.8} & \msd{88.5}{1.6} & \msd{88.1}{1.4} & \msd{\underline{89.6}}{0.4} & \msd{89.1}{0.7} & \msd{\bf 90.0}{\bf 0.3} \\
    \mymidline
    \multirow{3}{*}{\yelpfull} & 32 & \msd{30.2}{4.5} & \msd{32.7}{1.0} & \msd{32.9}{2.8} & \msd{36.8}{1.8} & \msd{35.2}{1.0} & \msd{32.6}{1.5} & \msd{35.9}{1.6} & \msd{\underline{40.1}}{2.2} & \msd{\bf 42.6}{\bf 1.1} \\
    & 64 & \msd{42.5}{1.7} & \msd{36.8}{2.1} & \msd{39.9}{3.4} & \msd{40.3}{2.6} & \msd{39.3}{\bf 1.0} & \msd{39.7}{1.8} & \msd{44.4}{1.1} & \msd{\underline{49.3}}{1.6} & \msd{\bf 49.8}{1.2} \\
    & 128 & \msd{47.7}{2.1} & \msd{41.3}{1.9} & \msd{49.4}{1.6} & \msd{45.1}{1.0} & \msd{46.4}{1.3} & \msd{46.8}{1.6} & \msd{\underline{51.2}}{0.8} & \msd{50.8}{1.5} & \msd{\bf 53.4}{\bf 0.7} \\
    \mymidline
    \multirow{3}{*}{\begin{tabular}[c]{@{}c@{}}AG's\\News\end{tabular}} & 32 & \msd{73.7}{4.6} & \msd{73.7}{3.0} & \msd{78.6}{1.6} & \msd{78.4}{2.3} & \msd{79.1}{2.7} & \msd{80.7}{1.8} & \msd{\underline{83.2}}{0.9} & \msd{81.8}{1.3} & \msd{\bf 83.7}{\bf 0.8} \\
    & 64 & \msd{80.0}{2.5} & \msd{80.0}{2.2} & \msd{82.0}{1.5} & \msd{82.6}{2.5} & \msd{82.4}{2.0} & \msd{83.0}{2.4} & \msd{\underline{85.3}}{0.7} & \msd{84.7}{1.3} & \msd{\bf 86.3}{\bf 0.6} \\
    & 128 & \msd{84.5}{1.7} & \msd{82.5}{0.8} & \msd{85.2}{0.6} & \msd{84.3}{1.7} & \msd{85.6}{0.8} & \msd{85.7}{0.3} & \msd{\underline{87.0}}{0.6} & \msd{86.2}{1.2} & \msd{\bf 87.5}{\bf 0.4} \\
    \mymidline
    \multirow{3}{*}{\begin{tabular}[c]{@{}c@{}}Yahoo!\\Answers\end{tabular}} & 32 & \msd{43.5}{4.0} & \msd{23.0}{1.6} & \msd{22.0}{2.3} & \msd{47.7}{2.3} & \msd{46.8}{2.1} & \msd{36.9}{1.8} & \msd{\underline{56.8}}{\bf 1.0} & \msd{54.5}{1.6} & \msd{\bf 58.0}{1.5} \\
    & 64 & \msd{53.1}{3.1} & \msd{37.6}{2.0} & \msd{45.7}{3.7} & \msd{55.3}{1.8} & \msd{52.9}{1.6} & \msd{54.0}{1.6} & \msd{\underline{61.9}}{\bf 0.7} & \msd{60.8}{1.4} & \msd{\bf 62.8}{1.3} \\
    & 128 & \msd{60.2}{1.5} & \msd{41.8}{1.9} & \msd{56.9}{2.5} & \msd{60.8}{1.9} & \msd{61.3}{1.0} & \msd{58.2}{1.5} & \msd{\underline{65.1}}{\bf 0.6} & \msd{64.3}{0.9} & \msd{\bf 66.2}{0.9} \\
    \mymidline
    \multirow{3}{*}{DBpedia} & 32 & \msd{67.1}{3.2} & \msd{18.9}{2.4} & \msd{64.0}{2.8} & \msd{77.5}{4.0} & \msd{83.3}{1.0} & \msd{78.2}{1.8} & \msd{\underline{85.3}}{\bf 0.9} & \msd{78.1}{2.6} & \msd{\bf 86.0}{1.7} \\
    & 64 & \msd{86.2}{2.4} & \msd{37.5}{3.0} & \msd{85.2}{0.8} & \msd{89.7}{1.1} & \msd{92.7}{0.9} & \msd{88.5}{0.7} & \msd{\underline{93.6}}{\bf 0.4} & \msd{92.7}{1.3} & \msd{\bf 94.1}{0.9} \\
    & 128 & \msd{95.0}{1.5} & \msd{47.5}{2.3} & \msd{89.4}{1.5} & \msd{95.7}{0.4} & \msd{96.5}{0.5} & \msd{95.7}{0.6} & \msd{\underline{97.0}}{\bf 0.2} & \msd{96.4}{0.4} & \msd{\bf 97.3}{0.3} \\
    \mymidline
    \multirow{3}{*}{TREC} & 32 & \msd{49.0}{3.5} & \msd{46.6}{1.4} & \msd{47.1}{3.6} & \msd{60.5}{3.7} & \msd{60.3}{1.5} & \msd{42.0}{4.4} & \msd{\underline{64.0}}{\bf 1.2} & \msd{57.6}{2.9} & \msd{\bf 70.2}{1.7} \\
    & 64 & \msd{69.1}{3.4} & \msd{59.8}{4.2} & \msd{75.7}{3.0} & \msd{73.0}{2.0} & \msd{77.3}{2.0} & \msd{72.6}{2.1} & \msd{78.6}{1.6} & \msd{\underline{81.8}}{3.1} & \msd{\bf 82.2}{\bf 1.5} \\
    & 128 & \msd{85.6}{2.8} & \msd{75.0}{1.8} & \msd{87.6}{3.0} & \msd{87.3}{3.6} & \msd{87.7}{1.5} & \msd{83.0}{3.8} & \msd{\underline{91.1}}{0.8} & \msd{89.7}{2.6} & \msd{\bf 92.1}{\bf 0.8} \\
    \mymidlines
    \multirow{3}{*}{\textit{Average}} & 32 & \msd{57.2}{3.8} & \msd{46.1}{2.1} & \msd{53.2}{2.7} & \msd{63.9}{3.0} & \msd{64.0}{1.8} & \msd{58.9}{2.5} & \msd{\underline{68.4}}{\bf 1.2} & \msd{66.3}{2.1} & \msd{\bf 71.2}{1.3} \\
    & 64 & \msd{68.9}{2.5} & \msd{56.1}{2.7} & \msd{68.5}{2.7} & \msd{71.2}{1.9} & \msd{71.6}{1.6} & \msd{70.3}{1.7} & \msd{75.2}{\bf 1.0} & \msd{\underline{76.2}}{1.8} & \msd{\bf 77.3}{1.1} \\
    & 128 & \msd{76.6}{1.9} & \msd{62.5}{1.7} & \msd{76.1}{1.9} & \msd{76.8}{1.9} & \msd{77.6}{1.2} & \msd{76.3}{1.9} & \msd{\underline{80.2}}{0.6} & \msd{79.4}{1.4} & \msd{\bf 81.1}{\bf 0.6} \\
    \bottomrule
\end{tabular}
    \end{adjustbox}
    \caption{Evaluation results of \mymethod{} and CSAL baselines on six datasets and three budgets (denoted by $b$), each with 10 repetitions.
    Accuracy (\%) of one-round fine-tuned PLM is reported in the format of \msd{\textrm{avg}}{\textrm{std}}.
    The \textbf{best} and \ul{second best} results per setup are {emboldened} and {underlined}, respectively.
    }
    \label{table:exp}
\end{table*}

\begin{figure}[t]
    \begin{subfigure}[t]{0.49\linewidth}
        \centering
        \includegraphics[width=\textwidth]{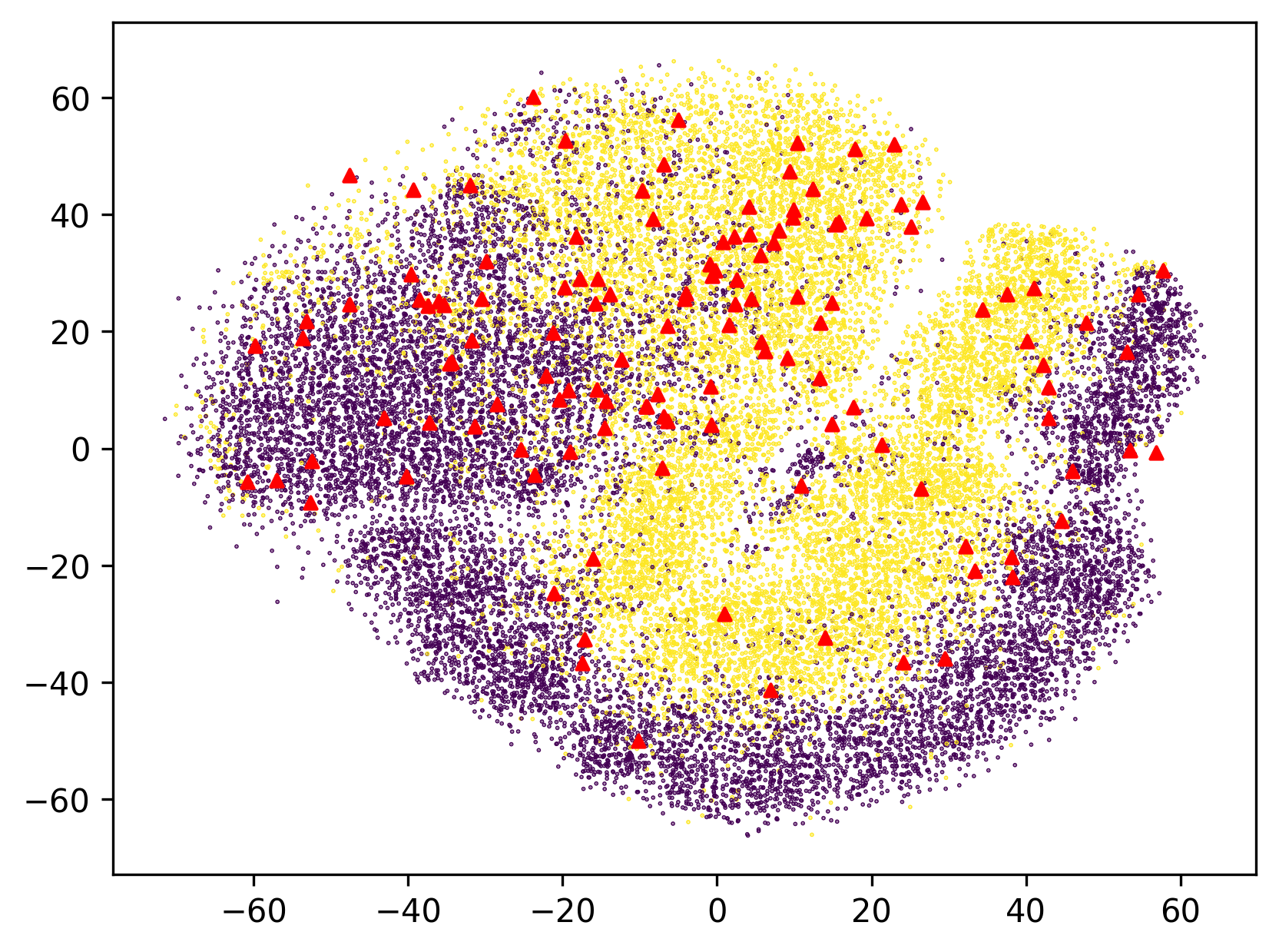}
        \caption{\votek{}.}
    \end{subfigure}
    \hfill
    \begin{subfigure}[t]{0.49\linewidth}
        \centering
        \includegraphics[width=\textwidth]{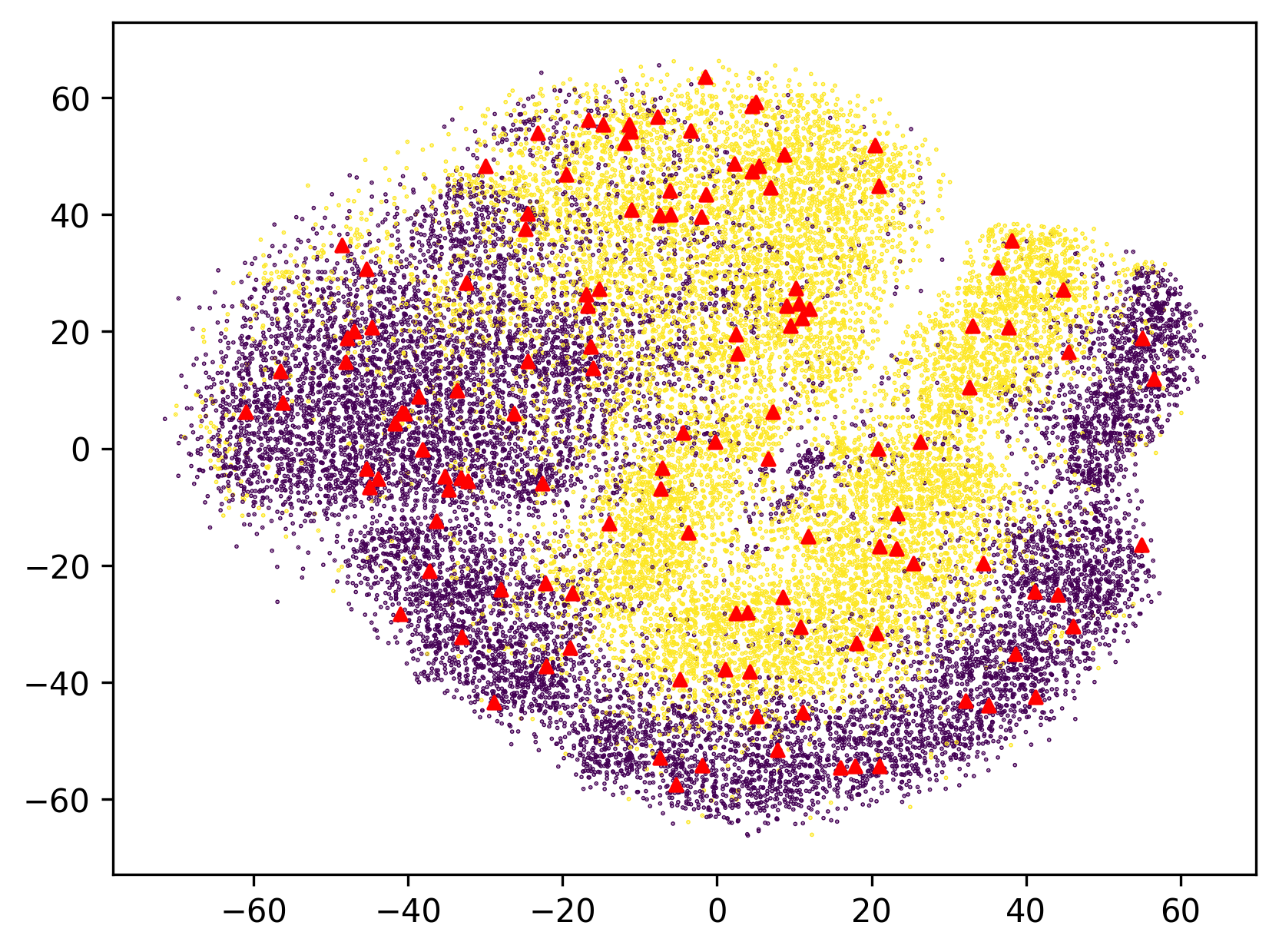}
        \caption{The proposed \mymethod{}.}
    \end{subfigure}
    \caption{The \tsne{} visualization of the acquired seed set ($b=128$) on IMDb dataset. Text embeddings are colored by their true labels.}
    \label{fig:tsne}
\end{figure}

The main quantitative results of PLM fine-tuning performance with \mymethod{} and baseline CSAL methods are shown in Table~\ref{table:exp}.
Results for baselines other than \votek{} are from \citet{yu-etal-2023-cold}.
To report the standard deviation, each setup is repeated with 10 different random seeds.
Figure~\ref{fig:tsne} demonstrates a qualitative visualization of the $b=128$ seed set from IMDb dataset, acquired by the latest baseline method \votek{} and the proposed \mymethod{}.
The \tsne{} \citep{tsne} method is used for visualization.

From results in Table~\ref{table:exp}, it can be seen that {\mymethod{}} consistently outperforms other baselines, achieving up to a $2.5\%$ gain on balanced datasets and up to $6.2\%$ on the imbalanced dataset, TREC.
{\mymethod{}} mainly benefits from that it enhances the class diversity as well as textual diversity. %
This can be concluded from the larger improvements on TREC.
In over half of the setups, \mymethod{} also achieves the lowest standard deviation.
In addition, \mymethod{} improves most when $b$ is small.
This aligns with the fundamental goal of AL, which is to maximize performance gains with minimal labeled data.
Furthermore, from the visualization in Figure~\ref{fig:tsne}, it can be seen that {\mymethod{}}'s enhancement of dual-diversity leads to a broader and more balanced coverage of both input space and label space. %
As \mymethod{} adopts a highest-uncertainty strategy, such coverage also exhibits high predictive uncertainty, thus including more ``hard examples'' which are valuable for annotation.

\begin{table*}[htb]
    \centering
    \begin{adjustbox}{width=\textwidth}
        \small
        \begin{tabular}{c|ccc|ccccc|c}
            \toprule
            \textbf{Dataset} & \textbf{Random} & \textbf{Entropy} & \textbf{Coreset} & \textbf{\alps} & \textbf{\textsc{few-s.}} & \textbf{TypiCl.} & \textbf{\patron} & \textbf{\votek} & \textbf{\mymethod{}} \\
            \mymidline
            IMDb & 1.207 & 6.111 & 1.000 & 1.783 & 1.286 & 2.765 & 1.286 & 1.065 & 1.169 \\
            \yelpfull & 1.778 & 3.800 & 6.000 & 2.833 & 2.000 & 5.200 & 2.250 & 1.273 & 1.450 \\
            AG's News & 1.462 & 28.000 & 2.000 & 1.667 & 1.500 & 1.818 & 1.500 & 2.200 & 1.133 \\
            Yahoo! Answers & 3.000 & 12.000 & 7.000 & 5.500 & 2.250 & 3.333 & 5.500 & 3.333 & 2.125 \\
            DBpedia & 3.500 & $\infty$ & 9.000 & 9.000 & 3.500 & 9.000 & 2.333 & 2.800 & 3.250 \\
            TREC & 8.000 & 16.000 & $\infty$ & 9.500 & 10.500 & 21.000 & 15.000 & 11.333 & 6.000 \\
            \mymidline
            \textit{Harmonic avg.} & 2.128 & 9.863 & 3.124 & 3.138 & 2.166 & 3.839 & 2.338 & \uline{2.052} & \textbf{1.779} \\
            \bottomrule
        \end{tabular}
    \end{adjustbox}
    \caption{Label imbalance value (IMB) of acquired seed sets ($b=128$). Smaller value indicates better class diversity and balance. An IMB of $\infty$ indicates that the \mse{} happens.
    }
    \label{table:analysis-imb}
\end{table*}

\begin{table*}[t]
    \centering
    \begin{adjustbox}{width=\textwidth}
        \small
        \begin{tabular}{c|ccc|ccccc|c}
            \toprule
            \textbf{Dataset} & \textbf{Random} & \textbf{Entropy} & \textbf{Coreset} & \textbf{\alps} & \textbf{\textsc{few-s.}} & \textbf{TypiCl.} & \textbf{\patron} & \textbf{\votek} & \textbf{\mymethod{}} \\
            \mymidline
            IMDb & 0.646 & 0.647 & 0.643 & 0.647 & 0.687 & 0.648 & 0.684 & 0.669 & 0.670 \\
            \yelpfull & 0.645 & 0.626 & 0.456 & 0.680 & 0.685 & 0.677 & 0.685 & 0.657 & 0.679 \\
            AG's News & 0.354 & 0.295 & 0.340 & 0.385 & 0.436 & 0.376 & 0.423 & 0.370 & 0.448 \\
            Yahoo! Answers & 0.430 & 0.375 & 0.400 & 0.441 & 0.470 & 0.438 & 0.486 & 0.451 & 0.491 \\
            DBpedia & 0.402 & 0.316 & 0.381 & 0.420 & 0.461 & 0.399 & 0.459 & 0.434 & 0.476 \\
            TREC & 0.301 & 0.298 & 0.298 & 0.339 & 0.337 & 0.326 & 0.338 & 0.346 & 0.353 \\
            \mymidline
            \textit{Average} & 0.463 & 0.426 & 0.420 & 0.485 & \uline{0.513} & 0.477 & 0.512 & 0.488 & \textbf{0.520} \\
            \bottomrule
        \end{tabular}
    \end{adjustbox}
    \caption{Textual diversity value $D$ of acquired seed sets ($b=128$). Larger values indicate better textual diversity.
    }
    \label{table:analysis-div}
\end{table*}

\subsection{Enhancement of Class Diversity}

To verify the enhancement of class diversity, the class imbalance value (IMB; \citealp{yu-etal-2023-cold}) under $b=128$ is reported in Table~\ref{table:analysis-imb}.

From Table~\ref{table:analysis-imb}, it can be seen that \mymethod{} achieves the lowest average IMB value.
This indicates that \mymethod{} enhances class diversity properly.
In contrast, an IMB of $\infty$ emerges in the pure uncertainty-based (Entropy) and textual-diversity-based (Coreset) method.
This indicates the \mse{} happens in their acquisition.

\subsection{Enhancement of Textual Diversity}

To measure the textual diversity of seed sets, the textual-diversity value \citep{ein-dor-etal-2020-active,yu-etal-2023-cold} under $b=128$ is reported in Table~\ref{table:analysis-div}.

Table~\ref{table:analysis-div} shows that \mymethod{} also achieves the highest average textual-diversity value.
This indicates that \mymethod{} also enhances textual diversity properly.
The improvement of textual-diversity value is not significant, compared to IMB value's (Table~\ref{table:analysis-imb}).
This signals that \mymethod{} enhances more of class diversity than textual diversity, compared to other baselines.
Such difference can be explained by the highest-uncertainty-candidate strategy, which acquires more information from the label space.

\subsection{Quality of Textual Embedding}

To analyze the quality of \mymethod{}'s prompt-based, \mbox{unsupervised} text embeddings $\tilde{\vect{z}}_{x_i}$ (\S\ref{subsec:prompt-pred-unc}),
they are compared with the supervised Sentence Transformer embeddings (\citealp{paraphrase-mpnet-base-v2}) used in \votek{} \citep{su2023selective}.
The correlations are computed across all the possible $\binom{N}{2}$ pairs of their cosine similarity.\footnote{Semantic similarity benchmarks (\eg{} STS) cannot be used here, as the prompt $T_x$ requires a task domain $\mathbb{Y}$.
    }
Results on three datasets are reported in Table~\ref{table:analysis-zx}.

\begin{table}[ht]
    \small
    \centering
    \begin{tabularx}{\linewidth}{X|cc}
        \toprule
        \textbf{Dataset} & \begin{tabular}[c]{@{}c@{}}\textbf{Pearson}\\\textbf{correlation $r$}\end{tabular} & \begin{tabular}[c]{@{}c@{}}\textbf{Spearman}\\\textbf{correlation $\rho$}\end{tabular} \\
        \mymidline
        IMDb & 0.1651 & 0.1636 \\
        {\small \quad{}\textit{w/ denoising}} & {\bf 0.1980} & {\bf 0.1889} \\
        \mymidline
        \yelpfull{} & 0.1424 & 0.1440 \\
        {\small \quad{}\textit{w/ denoising}} & {\bf 0.3072} & {\bf 0.2984} \\
        \mymidline
        TREC & 0.4271 & 0.4000 \\
        \small \quad{}\textit{w/ denoising} & \bf 0.4662 & \bf 0.4368 \\
        \bottomrule
    \end{tabularx}
    \caption{The quality of textual embeddings, \mbox{without} and with template denoising \citep{jiang-etal-2022-promptbert}. Both correlation metrics are over $[-1,1]$; higher values indicate better quality.
    }
    \label{table:analysis-zx}
\end{table}

From Table~\ref{table:analysis-zx}, a weak positive correlation is observed.
Moreover, template denoising produces better embeddings, as it removes the biases from raw embeddings.
Overall, the quality of textual embeddings is acceptable and adequate for cold-start acquisition.

\subsection{Quality of Class Prediction}

\begin{figure}[ht!]
    \includegraphics[width=\columnwidth]{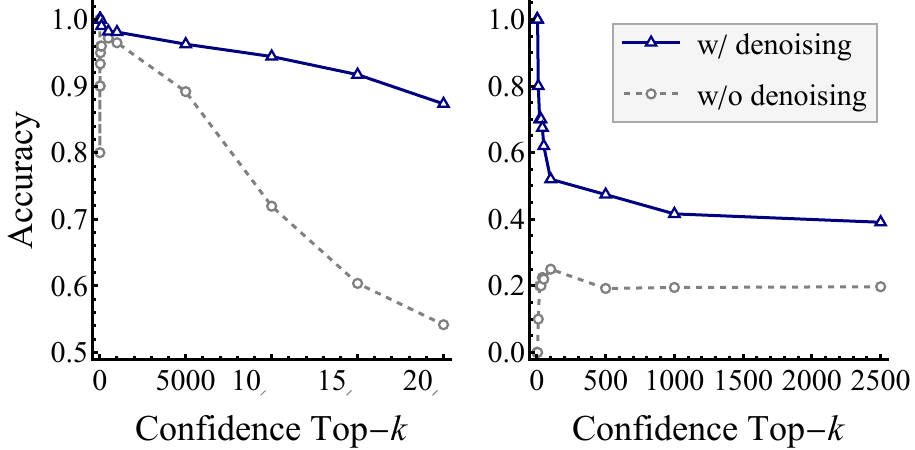}
    \caption{The quality of class predictions with respect to predictive uncertainty $u_i$.
    Dataset: IMDb (left) and TREC (right).
    }
    \label{fig:analysis-yhat}
\end{figure}

To analyze the quality of embedding-based class prediction $\hat{\vect{y}}_{i}$ (\S\ref{subsec:predict-module}),
they are compared with gold labels.
As uncertainty indicates unstable predictions, labels are arranged from the most confident (lowest $u_i$) to the least.
Results are demonstrated in Figure~\ref{fig:analysis-yhat}.

From Figure~\ref{fig:analysis-yhat}, a high accuracy of class predictions is consistently observed with high confidence and with denoised embeddings, and vice versa.
This demonstrates the good quality of e.d.f. predictions and the derived uncertainty metric.

\section{Discussion}

\subsection{Comparison with LLM-based Methods}

The landscape of NLP is rapidly evolving with generative large language models (LLMs).
This section evaluates two potential LLM-based alternatives to \mymethod{}: serialization for acquisition and zero-shot Chain-of-Thought prompting.
The following experiments are conducted with \llama{} \citep{touvron2023llama}.

\subsubsection{Serialization for Acquisition}

Inspired by the work of \citet{pmlr-v206-hegselmann23a}, class and uncertainty information can be serialized into natural language for LLM-based acquisition.
The process is designed to involve three passes.
In the first pass, each unlabeled text is formalized as a multiple-choice problem for LLM.
The prompt template $T_1$ is used to collect class and uncertainty information:
\begin{align*}
\begin{adjustbox}{max width=\columnwidth}
        $T_1 \coloneq \fbox{%
            \begin{varwidth}{\linewidth}%
                This sentence: ``\textph{}'' What is its \conph{}? \\
                Answer Choices: (A) [\textsc{class A}] (B) ... \\
                Answer: (
            \end{varwidth}%
        }$
\end{adjustbox}
\end{align*}
In the second pass, LLM decides on whether each text should be selected.
Predictive uncertainty is estimated by the entropy of first-pass predictions, bounded by $\log{C}$.
The extended template $T_2$ is used to combine multiple information:
\begin{align*}
\begin{adjustbox}{max width=\columnwidth}
    $T_2 \coloneq \fbox{%
        \begin{varwidth}{\linewidth}%
            This sentence: ``\textph{}'' What is its \conph{}? \\ Answer Choices: (A) [\textsc{class A}] (B) ... \\
            Answer: ([\textsc{answer}]) [\textsc{class}] \\
            Uncertainty: [\textsc{uncertainty} \%] \\
            Is it valuable for annotation? Yes or no? \\
            Answer:
        \end{varwidth}%
    }$
\end{adjustbox}
\end{align*}
In the third pass, texts with top-$b$ probabilities of $T_2$ answered ``yes'' are selected as the seed set.
LLM is then fine-tuned with the seed set under $T_1$.
Finally, $T_1$ is applied on the fine-tuned LLM to report the test set accuracy.

Due to resource constraints, LoRA (\citealp{hu2022lora}) is used for fine-tuning, with $r=\alpha=64$.
Results are reported in Table~\ref{table:llm-serial-main}.
Despite utilizing a mid-sized PLM, \mymethod{} outperforms serialization with LLM in most datasets.
The decision process of LLM is also black-box.
In contrast, \mymethod{} adopts graphs to explicitly capture the interplay of information, offering better interpretability.

\begin{table*}[t]
    \centering
    \begin{adjustbox}{}%
        \begin{tabular}{rc|cccccc!{\vrule width \lightrulewidth}c}
            \toprule
            \textbf{Method} & $b$ & \textbf{IMDb} & \textbf{\yelpfull} & \textbf{AG's News} & \textbf{Yahoo!} & \textbf{DBpedia} & \textbf{TREC} & \textit{\textbf{Average}} \\
            \midrule
            \multirow{3}{*}{Serialization} & 32 & 81.7 & \textbf{44.5} & 25.2 & 38.8 & 62.6 & 28.4 & 46.9 \\
            & 64 & 83.8 & \textbf{51.2} & 53.4 & 55.7 & 45.9 & 27.8 & 53.0 \\
            & 128 & 89.6 & \textbf{56.9} & 83.7 & 63.4 & 58.6 & 35.6 & 64.6 \\
            \midrule
            \multirow{3}{*}{\mymethod{}} & 32 & \textbf{86.9} & 42.6 & \textbf{83.7} & \textbf{58.0} & \textbf{86.0} & \textbf{70.2} & \textbf{71.2} \\
            & 64 & \textbf{88.5} & 49.8 & \textbf{86.3} & \textbf{62.8} & \textbf{94.1} & \textbf{82.2} & \textbf{77.3} \\
            & 128 & \textbf{90.0} & 53.4 & \textbf{87.5} & \textbf{66.2} & \textbf{97.3} & \textbf{92.1} & \textbf{81.1} \\
            \bottomrule
        \end{tabular}
    \end{adjustbox}
    \caption{
        Fine-tuning results of \mymethod{} (RoBERTa-base) and LLM serialization (\llama{}).
    }
    \label{table:llm-serial-main}
\end{table*}

\begin{table*}[ht]
    \centering
    \begin{adjustbox}{}%
        \begin{tabular}{r!{\vrule width \lightrulewidth}cccccc!{\vrule width \lightrulewidth}c}
            \toprule
            \textbf{Method} & \textbf{IMDb} & \textbf{\yelpfull} & \textbf{AG's News} & \textbf{Yahoo!} & \textbf{DBpedia} & \textbf{TREC} & \textbf{\textit{Average}} \\
            \midrule
            0-shot CoT, w/o choices & 63.6 & \hphantom{0}9.2 & 34.7 & 23.7 & 37.1 & 12.6 & 32.0 \\
            0-shot CoT, w/ choices & 72.1 & 25.4 & 60.2 & 43.6 & 32.3 & 24.2 & 43.0 \\
            \midrule
            \mymethod{}, $b=32$ & \textbf{86.9} & \textbf{42.6} & \textbf{83.7} & \textbf{58.0} & \textbf{86.0} & \textbf{70.2} & \textbf{71.2} \\
            \bottomrule
        \end{tabular}
    \end{adjustbox}
    \caption{
        Evaluation results of \mymethod{} ($b=32$, RoBERTa-base) and zero-shot Chain-of-Thought prompting (\citealp{NEURIPS2022_8bb0d291}; \llama{}).
    }
    \label{table:0shotCoT-no-choices-main}
\end{table*}

\begin{table}[t]
    \centering
    \begin{adjustbox}{max width=\columnwidth}
        \begin{tabular}{r!{\vrule width \lightrulewidth}cc!{\vrule width \lightrulewidth}cc}
            \toprule
            \multirow{2}{*}{\textbf{Stage}} & \multicolumn{2}{c!{\vrule width \lightrulewidth}}{\textbf{0-shot CoT }} & \multicolumn{2}{c}{\textbf{\mymethod{} }} \\
            & \begin{tabular}[c]{@{}c@{}}Energy\\(kJ)\end{tabular} & \begin{tabular}[c]{@{}c@{}}Time\\(sec)\end{tabular} & \begin{tabular}[c]{@{}c@{}}Energy\\(kJ)\end{tabular} & \begin{tabular}[c]{@{}c@{}}Time\\(sec)\end{tabular} \\
            \midrule
            Acquisition & \multicolumn{2}{c!{\vrule width \lightrulewidth}}{-} & \hphantom{0}59.82 & \hphantom{0}81.00 \\
            Fine-tuning & \multicolumn{2}{c!{\vrule width \lightrulewidth}}{-} & 225.77 & 208.89 \\
            Prediction & 2561.58 & 1967.23 & \hphantom{0}41.99 & \hphantom{0}24.27 \\
            \midrule
            \textit{Total} & 2561.58 & 1967.23 & \textbf{327.58} & \textbf{314.16} \\
            \bottomrule
        \end{tabular}
    \end{adjustbox}
    \caption{
        Energy consumption and time usage of \mymethod{} ($b=32$, RoBERTa-base) and zero-shot Chain-of-Thought prompting (\citealp{NEURIPS2022_8bb0d291}; \llama{}), under the same data amount of 25000.
    }
    \label{table:energy-main}
\end{table}

\begin{table*}[ht]
\centering
\begin{adjustbox}{width=\textwidth}%
    \begin{tabular}{rc!{\vrule width \lightrulewidth}cccccc!{\vrule width \lightrulewidth}c}
        \toprule
        \mymethod{} & $b$ & \textbf{IMDb} & \textbf{\yelpfull} & \textbf{AG's News} & \textbf{Yahoo!} & \textbf{DBpedia} & \textbf{TREC} & \textbf{\textit{Average}} \\
        \midrule
        \multirow{3}{*}{w/o noise} & 32 & \msd{86.9}{0.9} & \msd{42.6}{1.1} & \msd{83.7}{0.8} & \msd{58.0}{1.5} & \msd{86.0}{1.7} & \msd{70.2}{1.7} & \msd{71.2}{1.3} \\
        & 64 & \msd{88.5}{0.7} & \msd{49.8}{1.2} & \msd{86.3}{0.6} & \msd{62.8}{1.3} & \msd{94.1}{0.9} & \msd{82.2}{1.5} & \msd{77.2}{1.1} \\
        & 128 & \msd{90.0}{0.3} & \msd{53.4}{0.7} & \msd{87.5}{0.4} & \msd{66.2}{0.9} & \msd{97.3}{0.3} & \msd{92.1}{0.8} & \msd{81.1}{0.6} \\
        \midrule
        \multirow{3}{*}{w/ noise} & 32 & \msd{67.8}{4.3} & \msd{38.7}{3.0} & \msd{72.5}{1.0} & \msd{49.7}{7.2} & \msd{61.5}{2.0} & \msd{69.6}{0.6} & \msd{60.0}{1.5} \\
        & 64 & \msd{83.4}{1.3} & \msd{41.0}{2.7} & \msd{82.6}{1.4} & \msd{53.4}{2.7} & \msd{87.5}{3.3} & \msd{78.7}{3.3} & \msd{71.1}{1.1} \\
        & 128 & \msd{82.9}{6.3} & \msd{45.1}{1.7} & \msd{84.7}{2.4} & \msd{62.7}{1.3} & \msd{89.2}{3.7} & \msd{82.5}{3.8} & \msd{74.5}{1.5} \\
        \bottomrule
    \end{tabular}
\end{adjustbox}
\caption{
    Evaluation results of \mymethod{}, compared under an expected labeling noise level of 7\%.
}
\label{table:noisylabel-main}
\end{table*}

\begin{table*}[ht]
    \centering
    \begin{adjustbox}{width=\textwidth}%
        \begin{tabular}{rc!{\vrule width \lightrulewidth}cccccc!{\vrule width \lightrulewidth}c}
            \toprule
            & $b$ & \textbf{IMDb} & \textbf{\yelpfull} & \textbf{AG's News} & \textbf{Yahoo!} & \textbf{DBpedia} & \textbf{TREC} & \textbf{\textit{Average}} \\
            \midrule
            \multirow{3}{*}{Coreset} & 32 & \msd{74.5}{\bf2.9} & \msd{32.9}{2.8} & \msd{78.6}{\bf1.6} & \msd{22.0}{\bf2.3} & \msd{64.0}{2.8} & \msd{47.1}{\bf3.6} & \msd{53.2}{2.7} \\
            & 64 & \msd{82.8}{\bf2.5} & \msd{39.9}{3.4} & \msd{82.0}{1.5} & \msd{45.7}{3.7} & \msd{\bf 85.2}{\bf0.8} & \msd{75.7}{3.0} & \msd{68.5}{2.7} \\
            & 128 & \msd{87.8}{\bf0.8} & \msd{49.4}{1.6} & \msd{85.2}{0.6} & \msd{56.9}{2.5} & \msd{89.4}{1.5} & \msd{\bf 87.6}{3.0} & \msd{76.1}{1.9} \\
            \midrule
            \multirow{3}{*}{\begin{tabular}[c]{@{}r@{}}\mymethod{}\\w/ rand. pred.\end{tabular}} & 32 & \msd{\bf 83.3}{4.1} & \msd{\bf 44.1}{\bf0.7} & \msd{\bf 83.4}{2.0} & \msd{\bf 52.3}{3.9} & \msd{\bf 63.2}{\bf1.1} & \msd{\bf 64.9}{3.9} & \msd{\bf65.2}{\bf1.2} \\
            & 64 & \msd{\bf 85.9}{4.5} & \msd{\bf 48.0}{\bf0.3} & \msd{\bf 84.6}{\bf1.2} & \msd{\bf 60.0}{\bf0.6} & \msd{82.9}{1.7} & \msd{\bf 78.2}{\bf2.0} & \msd{\bf73.3}{\bf0.9} \\
            & 128 & \msd{\bf 86.6}{2.5} & \msd{\bf 49.5}{\bf0.4} & \msd{\bf 87.2}{\bf0.4} & \msd{\bf 63.4}{\bf1.3} & \msd{\bf 96.8}{\bf0.1} & \msd{86.8}{\bf1.3} & \msd{\bf78.4}{\bf0.5} \\
            \bottomrule
        \end{tabular}
    \end{adjustbox}
    \caption{
        Ablation results of \mymethod{} with random class predictions, compared with Coreset selection \citep{sener2018active}. In this case, the class and uncertainty information are disarranged.
    }
    \label{table:randomprediction-main}
\end{table*}

\subsubsection{Zero-shot Chain-of-Thought}

Zero-shot Chain-of-Thought (CoT) prompting \citep{NEURIPS2022_8bb0d291} with LLMs has emerged as a promising method in cold-start scenarios.
This paper tests zero-shot CoT without and with explicit choices in prompts.
The temperature of generation is set to 0, and a maximum of 256 tokens are generated.
Results are shown in Table~\ref{table:0shotCoT-no-choices-main}.
From the results, fine-tuning PLM with \mymethod{} still outperforms 0-shot LLM predictions.
In class-imbalanced and difficult datasets, performance gaps are greater.
Lemon-picking shows that the LLM failed to output a final answer within 256 tokens for many test instances.

In addition, the average total GPU and CPUs' energy consumption and time usage are measured using \citet{10.1145/3644815.3644967}'s method.
Results are reported in Table~\ref{table:energy-main}.
There is a 7.82$\times$ difference in energy consumption and 6.26$\times$ in time consumption.
While increasing the number of output tokens might improve, the added resource consumption cannot be neglected.
\mymethod{} provides an efficient solution for low-resource scenarios. %

\subsection{Effect of Labeling Noise}

Real-world annotations often involve noise.
\citet{BENCHMARKS2021_f2217062} estimated an average of 2.6\% labeling errors across 3 commonly-used NLP datasets.
To evaluate \mymethod{} under labeling noise, experiments with artificial errors are conducted.
As the gold labels may already contain around 3\% errors, 7\% of seed labels are randomly replaced by wrong labels.
The final sets are expected to exhibit an error level of 4--10\%.
Results are reported in Table~\ref{table:noisylabel-main}.

From the results, a decrease in accuracy and an increase in standard deviation occur as expected.
However, \mymethod{} still outperforms 0-shot CoT (Table~\ref{table:0shotCoT-no-choices-main}) in nearly all setups, despite the added noise.
This shows the robustness of \mymethod{} for fine-tuning to labeling noise.

\subsection{Effect of Class Prediction Failure}

For real-world cold-start tasks, the knowledge about classes might not be well exploited by the PLM.
In the worst case, the PLM can fail to generate meaningful class predictions.
To simulate this scenario, ablation experiments with random class predictions are conducted.
In this setup, the predictive embeddings $\vect{z}_{\predss}$ are replaced with random vectors.
This ablates class predictions.
Results are reported in Table~\ref{table:randomprediction-main}.

As class and uncertainty information are disarranged, \mymethod{} degenerates to single textual diversity and performance degradation occurs as expected.
Nonetheless, \mymethod{} still outperforms Coreset selection \citep{sener2018active}, a CSAL baseline which also purely utilizes textual diversity.
This demonstrates \mymethod{}'s effectiveness in real-world cold-start scenarios.

\subsection{Performance of Few-shot Math Reasoning}

\begin{table}[t]
    \centering
    \begin{adjustbox}{}%
        \begin{tabular}{r!{\vrule width \lightrulewidth}cc!{\vrule width \lightrulewidth}c}
            \toprule
            Method & 4-shot & 8-shot & \textit{Average} \\
            \midrule
            Random & 25.1 & 24.3 & {24.7} \\
            \mymethod{} & \textbf{25.8} & \textbf{27.4} & \textbf{26.6} \\
            \bottomrule
        \end{tabular}
    \end{adjustbox}
    \caption{
        Evaluation results of \mymethod{} (RoBERTa-base) with few-shot Chain-of-Thought prompting (\citealp{10.5555/3600270.3602070}; \llama{}) on GSM8K dataset \citep{cobbe2021gsm8k}, compared to random sampling.
    }
    \label{table:gsm8k-main}
\end{table}

\mymethod{} has the potential to generalize on other NLP tasks.
To demonstrate this, \mymethod{} is tested on GSM8K \citep{cobbe2021gsm8k}, a dataset of math word problems.
However, directly adapting RoBERTa to solving math problems is difficult due to its masked modeling nature.
Instead, \mymethod{} is applied with RoBERTa to produce a seed set.%
\footnote{
    For open questions like math problems, there are no concepts of ``classes''.
    Instead, the predictive embeddings $\tilde{\vect{z}}_{\predss}$ are clustered with \hdbscan{}.
    The cluster centroids are taken as \textit{meta}-class embeddings ${\vect{z}}_{\hat{y}}$.
}
Then, the seeds are taken as examples for few-shot Chain-of-Thought prompting \citep{10.5555/3600270.3602070} with \llama{}.
From the results, as reported in Table~\ref{table:gsm8k-main}, \mymethod{} is still effective in few-shot math problem solving, compared to random sampling.

\section{Conclusion}

This paper presents \mymethod{}, a dual-diversity enhancing and uncertainty-aware CSAL framework via a prompt-based and graph-based approach.
Different from previous works, it emphasizes dual-diversity (\ie{} textual diversity and class diversity) to ensure a balanced acquisition.
This is achieved by the novel construction of Dual-Neighbor Graph (DNG) and Farthest Point Sampling (FPS).
DNG leverages the $k$NN graph structure of textual space and label space from a PLM.
In addition, \mymethod{} prioritizes hard representative examples, so as to ensure an informative acquisition.
This leverages density-based clustering and uncertainty propagation on the DNG.
Experiments show the effectiveness of \mymethod{}'s dual-diversity enhancement and uncertainty-aware mechanism.
It offers an efficient solution for low-resource data acquisition.
Overall, \mymethod{}'s hybrid strategy strikes an important balance between exploration and exploitation in CSAL.

\section*{Limitations}

\paragraph{Backbone LM}
\mymethod{} leverages a discriminative PLM. %
However, state-of-the-art PLMs are primarily generative.
Generative embedding models (\eg{} \citealp{jiang2023scaling}) or adaptations \citep{10.5555/3454287.3454804,8983025,9369997} can be investigated and combined with \mymethod{}.
For such approaches, their quality and efficiency should be carefully minded.

\paragraph{External knowledge}
In \mymethod{}, the only source of external knowledge is the language model.
Incorporation of more domain knowledge, if possible, can improve the performance in the cold-start stage.
As \mymethod{} adopts a prompt-based and graph-based acquisition, prompt engineering and knowledge graphs \citep{kgllm} can be investigated.%

\section*{Acknowledgments}

We extend our gratitude to our action editor, Sebastian Padó, and the anonymous reviewers for their constructive comments.
We also thank Tianjun Li and Jiangfeng Liu for their helpful feedback on the initial drafts.

This work was funded in part by the National Natural Science Foundation of China grant under number 62222603, in part by the STI2030-Major Projects grant from the Ministry of Science and Technology of the People’s Republic of China under number 2021ZD0200700, in part by the Key-Area Research and Development Program of Guangdong Province under number 2023B0303030001, in part by the Program for Guangdong Introducing Innovative and Entrepreneurial Teams (2019ZT08X214), and in part by the Science and Technology Program of Guangzhou under number 2024A04J6310.

\bibliography{tacl2021}

\begin{thebibliography}{78}
\expandafter\ifx\csname natexlab\endcsname\relax\def\natexlab#1{#1}\fi

\bibitem[{Agarwal et~al.(2021)Agarwal, Srivastava, {Martin-del-Campo}, Natarajan, and Srinivasan}]{agarwal2021addressing}
Deepesh Agarwal, Pravesh Srivastava, Sergio {Martin-del-Campo}, Balasubramaniam Natarajan, and Babji Srinivasan. 2021.
\newblock \href {http://arxiv.org/abs/2110.03785v1} {Addressing practical challenges in active learning via a hybrid query strategy}.
\newblock \emph{arXiv preprint arXiv:2110.03785v1}.

\bibitem[{Aggarwal et~al.(2020)Aggarwal, Popescu, and Hudelot}]{9093475}
Umang Aggarwal, Adrian Popescu, and Céline Hudelot. 2020.
\newblock \href {https://doi.org/10.1109/WACV45572.2020.9093475} {Active learning for imbalanced datasets}.
\newblock In \emph{2020 IEEE Winter Conference on Applications of Computer Vision (WACV)}, pages 1417--1426.

\bibitem[{Alizadeh and Castor(2024)}]{10.1145/3644815.3644967}
Negar Alizadeh and Fernando Castor. 2024.
\newblock \href {https://doi.org/10.1145/3644815.3644967} {Green {AI}: A preliminary empirical study on energy consumption in {DL} models across different runtime infrastructures}.
\newblock In \emph{Proceedings of the IEEE/ACM 3rd International Conference on AI Engineering - Software Engineering for AI}, CAIN '24, pages 134--139, New York, NY, USA. Association for Computing Machinery.

\bibitem[{Ash et~al.(2020)Ash, Zhang, Krishnamurthy, Langford, and Agarwal}]{Ash2020Deep}
Jordan~T. Ash, Chicheng Zhang, Akshay Krishnamurthy, John Langford, and Alekh Agarwal. 2020.
\newblock \href {https://openreview.net/forum?id=ryghZJBKPS} {Deep batch active learning by diverse, uncertain gradient lower bounds}.
\newblock In \emph{8th International Conference on Learning Representations, {ICLR} 2020, Addis Ababa, Ethiopia, April 26--30, 2020}. OpenReview.net.

\bibitem[{Brangbour et~al.(2022)Brangbour, Bruneau, Tamisier, and Marchand-Maillet}]{brangbour2022cold}
Etienne Brangbour, Pierrick Bruneau, Thomas Tamisier, and St{\'e}phane Marchand-Maillet. 2022.
\newblock \href {http://arxiv.org/abs/2201.10227v1} {Cold start active learning strategies in the context of imbalanced classification}.
\newblock \emph{arXiv preprint arXiv:2201.10227v1}.

\bibitem[{Campello et~al.(2013)Campello, Moulavi, and Sander}]{10.1007/978-3-642-37456-2_14}
Ricardo J. G.~B. Campello, Davoud Moulavi, and Joerg Sander. 2013.
\newblock \href {https://doi.org/10.1007/978-3-642-37456-2_14} {Density-based clustering based on hierarchical density estimates}.
\newblock In \emph{Advances in Knowledge Discovery and Data Mining}, pages 160--172, Berlin, Heidelberg. Springer Berlin Heidelberg.

\bibitem[{Campello et~al.(2015)Campello, Moulavi, Zimek, and Sander}]{10.1145/2733381}
Ricardo J. G.~B. Campello, Davoud Moulavi, Arthur Zimek, and J\"{o}rg Sander. 2015.
\newblock \href {https://doi.org/10.1145/2733381} {Hierarchical density estimates for data clustering, visualization, and outlier detection}.
\newblock \emph{ACM Transactions on Knowledge Discovery from Data}, 10(1).

\bibitem[{Chang et~al.(2021)Chang, Shen, Yeh, and Demberg}]{chang-etal-2021-training}
Ernie Chang, Xiaoyu Shen, Hui-Syuan Yeh, and Vera Demberg. 2021.
\newblock \href {https://doi.org/10.18653/v1/2021.acl-short.2} {On training instance selection for few-shot neural text generation}.
\newblock In \emph{Proceedings of the 59th Annual Meeting of the Association for Computational Linguistics and the 11th International Joint Conference on Natural Language Processing (Volume 2: Short Papers)}, pages 8--13, Online. Association for Computational Linguistics.

\bibitem[{Cobbe et~al.(2021)Cobbe, Kosaraju, Bavarian, Chen, Jun, Kaiser, Plappert, Tworek, Hilton, Nakano, Hesse, and Schulman}]{cobbe2021gsm8k}
Karl Cobbe, Vineet Kosaraju, Mohammad Bavarian, Mark Chen, Heewoo Jun, Lukasz Kaiser, Matthias Plappert, Jerry Tworek, Jacob Hilton, Reiichiro Nakano, Christopher Hesse, and John Schulman. 2021.
\newblock \href {https://arxiv.org/abs/2110.14168v2} {Training verifiers to solve math word problems}.
\newblock \emph{arXiv preprint arXiv:2110.14168v2}.

\bibitem[{Dasgupta and Ng(2009)}]{dasgupta-ng-2009-mine}
Sajib Dasgupta and Vincent Ng. 2009.
\newblock \href {https://aclanthology.org/P09-1079} {Mine the easy, classify the hard: A semi-supervised approach to automatic sentiment classification}.
\newblock In \emph{Proceedings of the Joint Conference of the 47th Annual Meeting of the {ACL} and the 4th International Joint Conference on Natural Language Processing of the {AFNLP}}, pages 701--709, Suntec, Singapore. Association for Computational Linguistics.

\bibitem[{Dasgupta(2011)}]{DASGUPTA20111767}
Sanjoy Dasgupta. 2011.
\newblock \href {https://doi.org/10.1016/j.tcs.2010.12.054} {Two faces of active learning}.
\newblock \emph{Theoretical Computer Science}, 412(19):1767--1781.
\newblock Algorithmic Learning Theory (ALT 2009).

\bibitem[{De~Angeli et~al.(2021)De~Angeli, Gao, Alawad, Yoon, Schaefferkoetter, Wu, Durbin, Doherty, Stroup, Coyle, Penberthy, and Tourassi}]{DeAngeli2021}
Kevin De~Angeli, Shang Gao, Mohammed Alawad, Hong-Jun Yoon, Noah Schaefferkoetter, Xiao-Cheng Wu, Eric~B. Durbin, Jennifer Doherty, Antoinette Stroup, Linda Coyle, Lynne Penberthy, and Georgia Tourassi. 2021.
\newblock \href {https://doi.org/10.1186/s12859-021-04047-1} {Deep active learning for classifying cancer pathology reports}.
\newblock \emph{BMC Bioinformatics}, 22(1).

\bibitem[{Dligach and Palmer(2011)}]{dligach-palmer-2011-good}
Dmitriy Dligach and Martha Palmer. 2011.
\newblock \href {https://aclanthology.org/P11-2002} {Good seed makes a good crop: Accelerating active learning using language modeling}.
\newblock In \emph{Proceedings of the 49th Annual Meeting of the Association for Computational Linguistics: Human Language Technologies}, pages 6--10, Portland, Oregon, USA. Association for Computational Linguistics.

\bibitem[{Dubois and Prade(1982)}]{dubois:hal-04067331}
Didier Dubois and Henri Prade. 1982.
\newblock \href {https://doi.org/10.1080/03081078208934833} {A class of fuzzy measures based on triangular norms: A general framework for the combination of uncertain information}.
\newblock \emph{{International Journal of General Systems}}, 8(1):43--61.

\bibitem[{Ein-Dor et~al.(2020)Ein-Dor, Halfon, Gera, Shnarch, Dankin, Choshen, Danilevsky, Aharonov, Katz, and Slonim}]{ein-dor-etal-2020-active}
Liat Ein-Dor, Alon Halfon, Ariel Gera, Eyal Shnarch, Lena Dankin, Leshem Choshen, Marina Danilevsky, Ranit Aharonov, Yoav Katz, and Noam Slonim. 2020.
\newblock \href {https://doi.org/10.18653/v1/2020.emnlp-main.638} {Active learning for {BERT}: An empirical study}.
\newblock In \emph{Proceedings of the 2020 Conference on Empirical Methods in Natural Language Processing (EMNLP)}, pages 7949--7962, Online. Association for Computational Linguistics.

\bibitem[{Eklund and Forsman(2022)}]{eklund-forsman-2022-topic}
Anton Eklund and Mona Forsman. 2022.
\newblock \href {https://doi.org/10.18653/v1/2022.emnlp-industry.65} {Topic modeling by clustering language model embeddings: Human validation on an industry dataset}.
\newblock In \emph{Proceedings of the 2022 Conference on Empirical Methods in Natural Language Processing: Industry Track}, pages 635--643, Abu Dhabi, UAE. Association for Computational Linguistics.

\bibitem[{Eldar et~al.(1994)Eldar, Lindenbaum, Porat, and Zeevi}]{577129}
Yuval Eldar, Micahel Lindenbaum, Moshe Porat, and Yehoshua~Y. Zeevi. 1994.
\newblock \href {https://doi.org/10.1109/ICPR.1994.577129} {The farthest point strategy for progressive image sampling}.
\newblock In \emph{Proceedings of the 12th IAPR International Conference on Pattern Recognition, Vol. 2 - Conference B: Computer Vision \& Image Processing. (Cat. No.94CH3440-5)}, volume~3, pages 93--97.

\bibitem[{Fairstein et~al.(2024)Fairstein, Kalinsky, Karnin, Kushilevitz, Libov, and Tolmach}]{fairstein-etal-2024-class}
Yaron Fairstein, Oren Kalinsky, Zohar Karnin, Guy Kushilevitz, Alexander Libov, and Sofia Tolmach. 2024.
\newblock \href {https://aclanthology.org/2024.law-1.8} {Class balancing for efficient active learning in imbalanced datasets}.
\newblock In \emph{Proceedings of The 18th Linguistic Annotation Workshop (LAW-XVIII)}, pages 77--86, St. Julians, Malta. Association for Computational Linguistics.

\bibitem[{Gao et~al.(2019)Gao, He, Tan, Qin, Wang, and Liu}]{gao2018representation}
Jun Gao, Di~He, Xu~Tan, Tao Qin, Liwei Wang, and Tieyan Liu. 2019.
\newblock \href {https://openreview.net/forum?id=SkEYojRqtm} {Representation degeneration problem in training natural language generation models}.
\newblock In \emph{7th International Conference on Learning Representations, {ICLR} 2019, New Orleans, LA, USA, May 6--9, 2019}. OpenReview.net.

\bibitem[{Gao et~al.(2021)Gao, Yao, and Chen}]{gao-etal-2021-simcse}
Tianyu Gao, Xingcheng Yao, and Danqi Chen. 2021.
\newblock \href {https://doi.org/10.18653/v1/2021.emnlp-main.552} {{S}im{CSE}: Simple contrastive learning of sentence embeddings}.
\newblock In \emph{Proceedings of the 2021 Conference on Empirical Methods in Natural Language Processing}, pages 6894--6910, Online and Punta Cana, Dominican Republic. Association for Computational Linguistics.

\bibitem[{Gong et~al.(2019)Gong, Jin, and Zhang}]{8983025}
Xin-Rong Gong, Jian-Xiu Jin, and Tong Zhang. 2019.
\newblock \href {https://doi.org/10.1109/BIBM47256.2019.8983025} {Sentiment analysis using autoregressive language modeling and broad learning system}.
\newblock In \emph{2019 IEEE International Conference on Bioinformatics and Biomedicine (BIBM)}, pages 1130--1134.

\bibitem[{Hacohen et~al.(2022)Hacohen, Dekel, and Weinshall}]{pmlr-v162-hacohen22a}
Guy Hacohen, Avihu Dekel, and Daphna Weinshall. 2022.
\newblock \href {https://proceedings.mlr.press/v162/hacohen22a.html} {Active learning on a budget: Opposite strategies suit high and low budgets}.
\newblock In \emph{Proceedings of the 39th International Conference on Machine Learning}, volume 162 of \emph{Proceedings of Machine Learning Research}, pages 8175--8195. PMLR.

\bibitem[{Hegselmann et~al.(2023)Hegselmann, Buendia, Lang, Agrawal, Jiang, and Sontag}]{pmlr-v206-hegselmann23a}
Stefan Hegselmann, Alejandro Buendia, Hunter Lang, Monica Agrawal, Xiaoyi Jiang, and David Sontag. 2023.
\newblock \href {https://proceedings.mlr.press/v206/hegselmann23a.html} {{TabLLM}: Few-shot classification of tabular data with large language models}.
\newblock In \emph{Proceedings of The 26th International Conference on Artificial Intelligence and Statistics}, volume 206 of \emph{Proceedings of Machine Learning Research}, pages 5549--5581. PMLR.

\bibitem[{Herde et~al.(2021)Herde, Huseljic, Sick, and Calma}]{9650877}
Marek Herde, Denis Huseljic, Bernhard Sick, and Adrian Calma. 2021.
\newblock \href {https://doi.org/10.1109/ACCESS.2021.3135514} {A survey on cost types, interaction schemes, and annotator performance models in selection algorithms for active learning in classification}.
\newblock \emph{IEEE Access}, 9:166970--166989.

\bibitem[{Holzinger(2016)}]{Holzinger2016}
Andreas Holzinger. 2016.
\newblock \href {https://doi.org/10.1007/s40708-016-0042-6} {Interactive machine learning for health informatics: When do we need the human-in-the-loop?}
\newblock \emph{Brain Informatics}, 3(2):119--131.

\bibitem[{Hu et~al.(2022)Hu, Shen, Wallis, Allen{-}Zhu, Li, Wang, Wang, and Chen}]{hu2022lora}
Edward~J. Hu, Yelong Shen, Phillip Wallis, Zeyuan Allen{-}Zhu, Yuanzhi Li, Shean Wang, Lu~Wang, and Weizhu Chen. 2022.
\newblock \href {https://openreview.net/forum?id=nZeVKeeFYf9} {{L}o{RA}: Low-rank adaptation of large language models}.
\newblock In \emph{The Tenth International Conference on Learning Representations, {ICLR} 2022, Virtual Event, April 25--29, 2022}. OpenReview.net.

\bibitem[{Hu et~al.(2010)Hu, Namee, and Delany}]{hu2010off}
Rong Hu, Brian~Mac Namee, and Sarah~Jane Delany. 2010.
\newblock \href {http://www.aaai.org/ocs/index.php/FLAIRS/2010/paper/view/1305} {Off to a good start: Using clustering to select the initial training set in active learning}.
\newblock In \emph{Proceedings of the Twenty-Third International Florida Artificial Intelligence Research Society Conference, May 19--21, 2010, Daytona Beach, Florida, {USA}}. {AAAI} Press.

\bibitem[{Jiang et~al.(2023)Jiang, Huang, Luan, Wang, and Zhuang}]{jiang2023scaling}
Ting Jiang, Shaohan Huang, Zhongzhi Luan, Deqing Wang, and Fuzhen Zhuang. 2023.
\newblock \href {http://arxiv.org/abs/2307.16645v1} {Scaling sentence embeddings with large language models}.
\newblock \emph{arXiv preprint arXiv:2307.16645v1}.

\bibitem[{Jiang et~al.(2022)Jiang, Jiao, Huang, Zhang, Wang, Zhuang, Wei, Huang, Deng, and Zhang}]{jiang-etal-2022-promptbert}
Ting Jiang, Jian Jiao, Shaohan Huang, Zihan Zhang, Deqing Wang, Fuzhen Zhuang, Furu Wei, Haizhen Huang, Denvy Deng, and Qi~Zhang. 2022.
\newblock \href {https://doi.org/10.18653/v1/2022.emnlp-main.603} {{P}rompt{BERT}: Improving {BERT} sentence embeddings with prompts}.
\newblock In \emph{Proceedings of the 2022 Conference on Empirical Methods in Natural Language Processing}, pages 8826--8837, Abu Dhabi, United Arab Emirates. Association for Computational Linguistics.

\bibitem[{Jiang et~al.(2021)Jiang, Araki, Ding, and Neubig}]{jiang-etal-2021-know}
Zhengbao Jiang, Jun Araki, Haibo Ding, and Graham Neubig. 2021.
\newblock \href {https://doi.org/10.1162/tacl_a_00407} {How can we know when language models know? {O}n the calibration of language models for question answering}.
\newblock \emph{Transactions of the Association for Computational Linguistics}, 9:962--977.

\bibitem[{Kang et~al.(2004)Kang, Ryu, and Kwon}]{10.1007/978-3-540-24775-3_46}
Jaeho Kang, Kwang~Ryel Ryu, and Hyuk-Chul Kwon. 2004.
\newblock \href {https://doi.org/10.1007/978-3-540-24775-3_46} {Using cluster-based sampling to select initial training set for active learning in text classification}.
\newblock In \emph{Advances in Knowledge Discovery and Data Mining}, pages 384--388, Berlin, Heidelberg. Springer Berlin Heidelberg.

\bibitem[{Kojima et~al.(2022)Kojima, Gu, Reid, Matsuo, and Iwasawa}]{NEURIPS2022_8bb0d291}
Takeshi Kojima, Shixiang~Shane Gu, Machel Reid, Yutaka Matsuo, and Yusuke Iwasawa. 2022.
\newblock \href {https://proceedings.neurips.cc/paper_files/paper/2022/file/8bb0d291acd4acf06ef112099c16f326-Paper-Conference.pdf} {Large language models are zero-shot reasoners}.
\newblock In \emph{Advances in Neural Information Processing Systems}, volume~35, pages 22199--22213. Curran Associates, Inc.

\bibitem[{Krishnan et~al.(2021)Krishnan, Sinha, Ahuja, Subedar, Tickoo, and Iyer}]{sampling_bias_al}
Ranganath Krishnan, Alok Sinha, Nilesh~A. Ahuja, Mahesh Subedar, Omesh Tickoo, and Ravi~R. Iyer. 2021.
\newblock \href {https://icml.cc/virtual/2021/13299} {Mitigating sampling bias and improving robustness in active learning}.
\newblock In \emph{Proceedings of Workshop on Human in the Loop Learning (HILL) in International Conference on Machine Learning (ICML 2021)}.

\bibitem[{Lehmann et~al.(2015)Lehmann, Isele, Jakob, Jentzsch, Kontokostas, Mendes, Hellmann, Morsey, van Kleef, Auer, and Bizer}]{Lehmann2015}
Jens Lehmann, Robert Isele, Max Jakob, Anja Jentzsch, Dimitris Kontokostas, Pablo~N. Mendes, Sebastian Hellmann, Mohamed Morsey, Patrick van Kleef, S\"{o}ren Auer, and Christian Bizer. 2015.
\newblock \href {https://doi.org/10.3233/sw-140134} {{DB}pedia -- a large-scale, multilingual knowledge base extracted from {W}ikipedia}.
\newblock \emph{Semantic Web}, 6(2):167--195.

\bibitem[{Li and Roth(2002)}]{li-roth-2002-learning}
Xin Li and Dan Roth. 2002.
\newblock \href {https://aclanthology.org/C02-1150} {Learning question classifiers}.
\newblock In \emph{{COLING} 2002: The 19th International Conference on Computational Linguistics}.

\bibitem[{Li et~al.(2023)Li, Tan, and Liu}]{li2023privacy}
Yansong Li, Zhixing Tan, and Yang Liu. 2023.
\newblock \href {http://arxiv.org/abs/2305.06212v1} {Privacy-preserving prompt tuning for large language model services}.
\newblock \emph{arXiv preprint arXiv:2305.06212v1}.

\bibitem[{Liu et~al.(2019)Liu, Ott, Goyal, Du, Joshi, Chen, Levy, Lewis, Zettlemoyer, and Stoyanov}]{liu2019roberta}
Yinhan Liu, Myle Ott, Naman Goyal, Jingfei Du, Mandar Joshi, Danqi Chen, Omer Levy, Mike Lewis, Luke Zettlemoyer, and Veselin Stoyanov. 2019.
\newblock \href {http://arxiv.org/abs/1907.11692v1} {{RoBERTa}: A robustly optimized {BERT} pretraining approach}.
\newblock \emph{arXiv preprint arXiv:1907.11692v1}.

\bibitem[{Lu et~al.(2023)Lu, Yao, Zhang, Wang, Zhang, Lu, Li, and Wang}]{lu2023human}
Yuxuan Lu, Bingsheng Yao, Shao Zhang, Yun Wang, Peng Zhang, Tun Lu, Toby Jia-Jun Li, and Dakuo Wang. 2023.
\newblock \href {http://arxiv.org/abs/2311.09825v1} {Human still wins over {LLM}: An empirical study of active learning on domain-specific annotation tasks}.
\newblock \emph{arXiv preprint arXiv:2311.09825v1}.

\bibitem[{Maas et~al.(2011)Maas, Daly, Pham, Huang, Ng, and Potts}]{maas-etal-2011-learning}
Andrew~L. Maas, Raymond~E. Daly, Peter~T. Pham, Dan Huang, Andrew~Y. Ng, and Christopher Potts. 2011.
\newblock \href {https://aclanthology.org/P11-1015} {Learning word vectors for sentiment analysis}.
\newblock In \emph{Proceedings of the 49th Annual Meeting of the Association for Computational Linguistics: Human Language Technologies}, pages 142--150, Portland, Oregon, USA. Association for Computational Linguistics.

\bibitem[{van~der Maaten and Hinton(2008)}]{tsne}
Laurens van~der Maaten and Geoffrey Hinton. 2008.
\newblock \href {http://jmlr.org/papers/v9/vandermaaten08a.html} {Visualizing data using t-{SNE}}.
\newblock \emph{Journal of Machine Learning Research}, 9(86):2579--2605.

\bibitem[{Marcheggiani and Arti{\`e}res(2014)}]{marcheggiani-artieres-2014-experimental}
Diego Marcheggiani and Thierry Arti{\`e}res. 2014.
\newblock \href {https://doi.org/10.3115/v1/D14-1097} {An experimental comparison of active learning strategies for partially labeled sequences}.
\newblock In \emph{Proceedings of the 2014 Conference on Empirical Methods in Natural Language Processing ({EMNLP})}, pages 898--906, Doha, Qatar. Association for Computational Linguistics.

\bibitem[{Margatina et~al.(2021)Margatina, Vernikos, Barrault, and Aletras}]{margatina-etal-2021-active}
Katerina Margatina, Giorgos Vernikos, Lo{\"\i}c Barrault, and Nikolaos Aletras. 2021.
\newblock \href {https://doi.org/10.18653/v1/2021.emnlp-main.51} {Active learning by acquiring contrastive examples}.
\newblock In \emph{Proceedings of the 2021 Conference on Empirical Methods in Natural Language Processing}, pages 650--663, Online and Punta Cana, Dominican Republic. Association for Computational Linguistics.

\bibitem[{McInnes et~al.(2020)McInnes, Healy, and Melville}]{mcinnes2018umap}
Leland McInnes, John Healy, and James Melville. 2020.
\newblock \href {http://arxiv.org/abs/1802.03426v3} {{UMAP}: Uniform manifold approximation and projection for dimension reduction}.
\newblock \emph{arXiv preprint arXiv:1802.03426v3}.

\bibitem[{McInnes et~al.(2018)McInnes, Healy, Saul, and Großberger}]{McInnes2018}
Leland McInnes, John Healy, Nathaniel Saul, and Lukas Großberger. 2018.
\newblock \href {https://doi.org/10.21105/joss.00861} {\texttt{UMAP}: Uniform manifold approximation and projection}.
\newblock \emph{Journal of Open Source Software}, 3(29):861.

\bibitem[{Meng et~al.(2019)Meng, Shen, Zhang, and Han}]{Meng_Shen_Zhang_Han_2019}
Yu~Meng, Jiaming Shen, Chao Zhang, and Jiawei Han. 2019.
\newblock \href {https://doi.org/10.1609/aaai.v33i01.33016826} {Weakly-supervised hierarchical text classification}.
\newblock \emph{Proceedings of the AAAI Conference on Artificial Intelligence}, 33(01):6826--6833.

\bibitem[{Miao et~al.(2023)Miao, Du, and Zhang}]{10.1145/3583780.3614833}
Pu~Miao, Zeyao Du, and Junlin Zhang. 2023.
\newblock \href {https://doi.org/10.1145/3583780.3614833} {{DebCSE}: Rethinking unsupervised contrastive sentence embedding learning in the debiasing perspective}.
\newblock In \emph{Proceedings of the 32nd ACM International Conference on Information and Knowledge Management}, CIKM '23, pages 1847--1856, New York, NY, USA. Association for Computing Machinery.

\bibitem[{M\"{u}ller et~al.(2022)M\"{u}ller, P\'{e}rez-Torr\'{o}, Basile, and Franco-Salvador}]{10.1007/978-3-031-08473-7_9}
Thomas M\"{u}ller, Guillermo P\'{e}rez-Torr\'{o}, Angelo Basile, and Marc Franco-Salvador. 2022.
\newblock \href {https://doi.org/10.1007/978-3-031-08473-7_9} {Active few-shot learning with {FASL}}.
\newblock In \emph{Natural Language Processing and Information Systems; 27th International Conference on Applications of Natural Language to Information Systems, NLDB 2022, Valencia, Spain, June 15--17, 2022, Proceedings}, pages 98--110, Cham. Springer International Publishing.

\bibitem[{Naeini et~al.(2023)Naeini, Saqur, Saeidi, Giorgi, and Taati}]{naeini2023large}
Saeid~Alavi Naeini, Raeid Saqur, Mozhgan Saeidi, John~Michael Giorgi, and Babak Taati. 2023.
\newblock \href {https://openreview.net/forum?id=ZV4tZgclu8} {Large language models are fixated by red herrings: Exploring creative problem solving and {E}instellung effect using the {O}nly {C}onnect {W}all dataset}.
\newblock In \emph{Thirty-seventh Conference on Neural Information Processing Systems Datasets and Benchmarks Track}.

\bibitem[{Nguyen and Smeulders(2004)}]{10.1145/1015330.1015349}
Hieu~T. Nguyen and Arnold Smeulders. 2004.
\newblock \href {https://doi.org/10.1145/1015330.1015349} {Active learning using pre-clustering}.
\newblock In \emph{Proceedings of the Twenty-First International Conference on Machine Learning}, ICML '04, page~79, New York, NY, USA. Association for Computing Machinery.

\bibitem[{Northcutt et~al.(2021)Northcutt, Athalye, and Mueller}]{BENCHMARKS2021_f2217062}
Curtis Northcutt, Anish Athalye, and Jonas Mueller. 2021.
\newblock \href {https://datasets-benchmarks-proceedings.neurips.cc/paper_files/paper/2021/file/f2217062e9a397a1dca429e7d70bc6ca-Paper-round1.pdf} {Pervasive label errors in test sets destabilize machine learning benchmarks}.
\newblock In \emph{Proceedings of the Neural Information Processing Systems Track on Datasets and Benchmarks}, volume~1.

\bibitem[{Padhy et~al.(2020)Padhy, Nado, Ren, Liu, Snoek, and Lakshminarayanan}]{ova-unc}
Shreyas Padhy, Zachary Nado, Jie Ren, Jeremiah Liu, Jasper Snoek, and Balaji Lakshminarayanan. 2020.
\newblock \href {https://www.gatsby.ucl.ac.uk/~balaji/udl2020/accepted-papers/UDL2020-paper-040.pdf} {Revisiting one-vs-all classifiers for predictive uncertainty and out-of-distribution detection in neural networks}.
\newblock In \emph{ICML 2020 Workshop on Uncertainty and Robustness in Deep Learning}.

\bibitem[{Pan et~al.(2024)Pan, Luo, Wang, Chen, Wang, and Wu}]{kgllm}
Shirui Pan, Linhao Luo, Yufei Wang, Chen Chen, Jiapu Wang, and Xindong Wu. 2024.
\newblock \href {https://doi.org/10.1109/TKDE.2024.3352100} {Unifying large language models and knowledge graphs: A roadmap}.
\newblock \emph{IEEE Transactions on Knowledge and Data Engineering}, pages 1--20.

\bibitem[{Park and Caragea(2022)}]{park-caragea-2022-calibration}
Seo~Yeon Park and Cornelia Caragea. 2022.
\newblock \href {https://doi.org/10.18653/v1/2022.acl-long.368} {On the calibration of pre-trained language models using mixup guided by area under the margin and saliency}.
\newblock In \emph{Proceedings of the 60th Annual Meeting of the Association for Computational Linguistics (Volume 1: Long Papers)}, pages 5364--5374, Dublin, Ireland. Association for Computational Linguistics.

\bibitem[{Rudolph et~al.(2023)Rudolph, Williams, Miles, Antonelli, and Diaz}]{random_instability}
Kara~E. Rudolph, Nicholas~T. Williams, Caleb~H. Miles, Joseph Antonelli, and Ivan Diaz. 2023.
\newblock \href {https://doi.org/doi:10.1515/jci-2023-0022} {All models are wrong, but which are useful? {C}omparing parametric and nonparametric estimation of causal effects in finite samples}.
\newblock \emph{Journal of Causal Inference}, 11(1):20230022.

\bibitem[{Schr{\"o}der et~al.(2022)Schr{\"o}der, Niekler, and Potthast}]{schroder-etal-2022-revisiting}
Christopher Schr{\"o}der, Andreas Niekler, and Martin Potthast. 2022.
\newblock \href {https://doi.org/10.18653/v1/2022.findings-acl.172} {Revisiting uncertainty-based query strategies for active learning with transformers}.
\newblock In \emph{Findings of the Association for Computational Linguistics: ACL 2022}, pages 2194--2203, Dublin, Ireland. Association for Computational Linguistics.

\bibitem[{Sch\"{u}tze et~al.(2006)Sch\"{u}tze, Velipasaoglu, and Pedersen}]{10.1145/1183614.1183709}
Hinrich Sch\"{u}tze, Emre Velipasaoglu, and Jan~O. Pedersen. 2006.
\newblock \href {https://doi.org/10.1145/1183614.1183709} {Performance thresholding in practical text classification}.
\newblock In \emph{Proceedings of the 15th ACM International Conference on Information and Knowledge Management}, CIKM '06, pages 662--671, New York, NY, USA. Association for Computing Machinery.

\bibitem[{Sener and Savarese(2018)}]{sener2018active}
Ozan Sener and Silvio Savarese. 2018.
\newblock \href {https://openreview.net/forum?id=H1aIuk-RW} {Active learning for convolutional neural networks: {A} core-set approach}.
\newblock In \emph{6th International Conference on Learning Representations, {ICLR} 2018, Vancouver, BC, Canada, April 30 -- May 3, 2018, Conference Track Proceedings}. OpenReview.net.

\bibitem[{{Sentence Transformers}(2024)}]{paraphrase-mpnet-base-v2}
{Sentence Transformers}. 2024.
\newblock \href {https://doi.org/10.57967/hf/2004} {\texttt{paraphrase-mpnet-base-v2} (revision \texttt{e6981e5})}.
\newblock Hugging Face.

\bibitem[{Settles(2009)}]{settles.tr09}
Burr Settles. 2009.
\newblock \href {https://research.cs.wisc.edu/techreports/2009/TR1648.pdf} {Active learning literature survey}.
\newblock Computer Sciences Technical Report 1648, University of Wisconsin--Madison.

\bibitem[{Shnarch et~al.(2022)Shnarch, Gera, Halfon, Dankin, Choshen, Aharonov, and Slonim}]{shnarch-etal-2022-cluster}
Eyal Shnarch, Ariel Gera, Alon Halfon, Lena Dankin, Leshem Choshen, Ranit Aharonov, and Noam Slonim. 2022.
\newblock \href {https://doi.org/10.18653/v1/2022.acl-long.526} {Cluster {\&} tune: {B}oost cold start performance in text classification}.
\newblock In \emph{Proceedings of the 60th Annual Meeting of the Association for Computational Linguistics (Volume 1: Long Papers)}, pages 7639--7653, Dublin, Ireland. Association for Computational Linguistics.

\bibitem[{Su et~al.(2023)Su, Kasai, Wu, Shi, Wang, Xin, Zhang, Ostendorf, Zettlemoyer, Smith, and Yu}]{su2023selective}
Hongjin Su, Jungo Kasai, Chen~Henry Wu, Weijia Shi, Tianlu Wang, Jiayi Xin, Rui Zhang, Mari Ostendorf, Luke Zettlemoyer, Noah~A. Smith, and Tao Yu. 2023.
\newblock \href {https://openreview.net/forum?id=qY1hlv7gwg} {Selective annotation makes language models better few-shot learners}.
\newblock In \emph{The Eleventh International Conference on Learning Representations, {ICLR} 2023, Kigali, Rwanda, May 1--5, 2023}. OpenReview.net.

\bibitem[{Tomanek et~al.(2009)Tomanek, Laws, Hahn, and Sch{\"u}tze}]{tomanek-etal-2009-proper}
Katrin Tomanek, Florian Laws, Udo Hahn, and Hinrich Sch{\"u}tze. 2009.
\newblock \href {https://aclanthology.org/W09-1902} {On proper unit selection in active learning: Co-selection effects for named entity recognition}.
\newblock In \emph{Proceedings of the {NAACL} {HLT} 2009 Workshop on Active Learning for Natural Language Processing}, pages 9--17, Boulder, Colorado. Association for Computational Linguistics.

\bibitem[{Touvron et~al.(2023)Touvron, Martin, Stone, Albert, Almahairi, Babaei, Bashlykov, Batra, Bhargava, Bhosale, Bikel, Blecher, Ferrer, Chen, Cucurull, Esiobu, Fernandes, Fu, Fu, Fuller, Gao, Goswami, Goyal, Hartshorn, Hosseini, Hou, Inan, Kardas, Kerkez, Khabsa, Kloumann, Korenev, Koura, Lachaux, Lavril, Lee, Liskovich, Lu, Mao, Martinet, Mihaylov, Mishra, Molybog, Nie, Poulton, Reizenstein, Rungta, Saladi, Schelten, Silva, Smith, Subramanian, Tan, Tang, Taylor, Williams, Kuan, Xu, Yan, Zarov, Zhang, Fan, Kambadur, Narang, Rodriguez, Stojnic, Edunov, and Scialom}]{touvron2023llama}
Hugo Touvron, Louis Martin, Kevin Stone, Peter Albert, Amjad Almahairi, Yasmine Babaei, Nikolay Bashlykov, Soumya Batra, Prajjwal Bhargava, Shruti Bhosale, Dan Bikel, Lukas Blecher, Cristian~Canton Ferrer, Moya Chen, Guillem Cucurull, David Esiobu, Jude Fernandes, Jeremy Fu, Wenyin Fu, Brian Fuller, Cynthia Gao, Vedanuj Goswami, Naman Goyal, Anthony Hartshorn, Saghar Hosseini, Rui Hou, Hakan Inan, Marcin Kardas, Viktor Kerkez, Madian Khabsa, Isabel Kloumann, Artem Korenev, Punit~Singh Koura, Marie-Anne Lachaux, Thibaut Lavril, Jenya Lee, Diana Liskovich, Yinghai Lu, Yuning Mao, Xavier Martinet, Todor Mihaylov, Pushkar Mishra, Igor Molybog, Yixin Nie, Andrew Poulton, Jeremy Reizenstein, Rashi Rungta, Kalyan Saladi, Alan Schelten, Ruan Silva, Eric~Michael Smith, Ranjan Subramanian, Xiaoqing~Ellen Tan, Binh Tang, Ross Taylor, Adina Williams, Jian~Xiang Kuan, Puxin Xu, Zheng Yan, Iliyan Zarov, Yuchen Zhang, Angela Fan, Melanie Kambadur, Sharan Narang, Aurelien Rodriguez, Robert Stojnic, Sergey Edunov, and Thomas Scialom. 2023.
\newblock \href {http://arxiv.org/abs/2307.09288v2} {Llama 2: Open foundation and fine-tuned chat models}.
\newblock \emph{arXiv preprint arXiv:2307.09288v2}.

\bibitem[{Wang(2024)}]{wang2023calibration}
Cheng Wang. 2024.
\newblock \href {http://arxiv.org/abs/2308.01222v2} {Calibration in deep learning: A survey of the state-of-the-art}.
\newblock \emph{arXiv preprint arXiv:2308.01222v2}.

\bibitem[{Wei et~al.(2022)Wei, Wang, Schuurmans, Bosma, Ichter, Xia, Chi, Le, and Zhou}]{10.5555/3600270.3602070}
Jason Wei, Xuezhi Wang, Dale Schuurmans, Maarten Bosma, Brian Ichter, Fei Xia, Ed~Chi, Quoc~V. Le, and Denny Zhou. 2022.
\newblock \href {https://proceedings.neurips.cc/paper_files/paper/2022/file/9d5609613524ecf4f15af0f7b31abca4-Paper-Conference.pdf} {Chain-of-thought prompting elicits reasoning in large language models}.
\newblock In \emph{Advances in Neural Information Processing Systems}, volume~35, pages 24824--24837. Curran Associates, Inc.

\bibitem[{Wu et~al.(2022)Wu, Xiao, Sun, Zhang, Ma, and He}]{WU2022364}
Xingjiao Wu, Luwei Xiao, Yixuan Sun, Junhang Zhang, Tianlong Ma, and Liang He. 2022.
\newblock \href {https://doi.org/10.1016/j.future.2022.05.014} {A survey of human-in-the-loop for machine learning}.
\newblock \emph{Future Generation Computer Systems}, 135:364--381.

\bibitem[{Wójcik et~al.(2022)Wójcik, Grela, Śmieja, Misztal, and Tabor}]{WOJCIK2022109219}
Bartosz Wójcik, Jacek Grela, Marek Śmieja, Krzysztof Misztal, and Jacek Tabor. 2022.
\newblock \href {https://doi.org/10.1016/j.asoc.2022.109219} {{SLOVA}: Uncertainty estimation using single label one-vs-all classifier}.
\newblock \emph{Applied Soft Computing}, 126:109219.

\bibitem[{Yang et~al.(2019)Yang, Dai, Yang, Carbonell, Salakhutdinov, and Le}]{10.5555/3454287.3454804}
Zhilin Yang, Zihang Dai, Yiming Yang, Jaime Carbonell, Russ~R. Salakhutdinov, and Quoc~V. Le. 2019.
\newblock \href {https://proceedings.neurips.cc/paper_files/paper/2019/file/dc6a7e655d7e5840e66733e9ee67cc69-Paper.pdf} {{XLN}et: Generalized autoregressive pretraining for language understanding}.
\newblock In \emph{Advances in Neural Information Processing Systems}, volume~32. Curran Associates, Inc.

\bibitem[{Yu et~al.(2019)Yu, Yang, Zheng, and Sun}]{8443399}
Hualong Yu, Xibei Yang, Shang Zheng, and Changyin Sun. 2019.
\newblock \href {https://doi.org/10.1109/TNNLS.2018.2855446} {Active learning from imbalanced data: A solution of online weighted extreme learning machine}.
\newblock \emph{IEEE Transactions on Neural Networks and Learning Systems}, 30(4):1088--1103.

\bibitem[{Yu et~al.(2023)Yu, Zhang, Xu, Zhang, Shen, and Zhang}]{yu-etal-2023-cold}
Yue Yu, Rongzhi Zhang, Ran Xu, Jieyu Zhang, Jiaming Shen, and Chao Zhang. 2023.
\newblock \href {https://doi.org/10.18653/v1/2023.acl-long.141} {Cold-start data selection for better few-shot language model fine-tuning: A prompt-based uncertainty propagation approach}.
\newblock In \emph{Proceedings of the 61st Annual Meeting of the Association for Computational Linguistics (Volume 1: Long Papers)}, pages 2499--2521, Toronto, Canada. Association for Computational Linguistics.

\bibitem[{Yuan et~al.(2020{\natexlab{a}})Yuan, Lin, and Boyd-Graber}]{yuan-etal-2020-cold}
Michelle Yuan, Hsuan-Tien Lin, and Jordan Boyd-Graber. 2020{\natexlab{a}}.
\newblock \href {https://doi.org/10.18653/v1/2020.emnlp-main.637} {Cold-start active learning through self-supervised language modeling}.
\newblock In \emph{Proceedings of the 2020 Conference on Empirical Methods in Natural Language Processing (EMNLP)}, pages 7935--7948, Online. Association for Computational Linguistics.

\bibitem[{Yuan et~al.(2020{\natexlab{b}})Yuan, Zhang, Li, and Xiong}]{9101545}
Mu~Yuan, Lan Zhang, Xiang-Yang Li, and Hui Xiong. 2020{\natexlab{b}}.
\newblock \href {https://doi.org/10.1109/ICDE48307.2020.00188} {Comprehensive and efficient data labeling via adaptive model scheduling}.
\newblock In \emph{2020 IEEE 36th International Conference on Data Engineering (ICDE)}, pages 1858--1861.

\bibitem[{Zhang et~al.(2023)Zhang, Wu, and Zhang}]{zhang2023utilising}
Shiwei Zhang, Mingfang Wu, and Xiuzhen Zhang. 2023.
\newblock \href {http://arxiv.org/abs/2310.11318v1} {Utilising a large language model to annotate subject metadata: A case study in an {A}ustralian national research data catalogue}.
\newblock \emph{arXiv preprint arXiv:2310.11318v1}.

\bibitem[{Zhang et~al.(2022{\natexlab{a}})Zhang, Gong, and Chen}]{9369997}
Tong Zhang, Xinrong Gong, and C.~L.~Philip Chen. 2022{\natexlab{a}}.
\newblock \href {https://doi.org/10.1109/TCYB.2021.3050508} {{BMT-N}et: Broad multitask transformer network for sentiment analysis}.
\newblock \emph{IEEE Transactions on Cybernetics}, 52(7):6232--6243.

\bibitem[{Zhang et~al.(2021)Zhang, Su, Qing, Xu, Cai, and Xing}]{8654015}
Tong Zhang, Guoxi Su, Chunmei Qing, Xiangmin Xu, Bolun Cai, and Xiaofen Xing. 2021.
\newblock \href {https://doi.org/10.1109/TSMC.2018.2884996} {Hierarchical lifelong learning by sharing representations and integrating hypothesis}.
\newblock \emph{IEEE Transactions on Systems, Man, and Cybernetics: Systems}, 51(2):1004--1014.

\bibitem[{Zhang et~al.(2015)Zhang, Zhao, and LeCun}]{NIPS2015_250cf8b5}
Xiang Zhang, Junbo Zhao, and Yann LeCun. 2015.
\newblock \href {https://proceedings.neurips.cc/paper_files/paper/2015/file/250cf8b51c773f3f8dc8b4be867a9a02-Paper.pdf} {Character-level convolutional networks for text classification}.
\newblock In \emph{Advances in Neural Information Processing Systems}, volume~28. Curran Associates, Inc.

\bibitem[{Zhang et~al.(2022{\natexlab{b}})Zhang, Strubell, and Hovy}]{zhang-etal-2022-survey}
Zhisong Zhang, Emma Strubell, and Eduard Hovy. 2022{\natexlab{b}}.
\newblock \href {https://doi.org/10.18653/v1/2022.emnlp-main.414} {A survey of active learning for natural language processing}.
\newblock In \emph{Proceedings of the 2022 Conference on Empirical Methods in Natural Language Processing}, pages 6166--6190, Abu Dhabi, United Arab Emirates. Association for Computational Linguistics.

\bibitem[{Zhu et~al.(2008)Zhu, Wang, Yao, and Tsou}]{zhu-etal-2008-active}
Jingbo Zhu, Huizhen Wang, Tianshun Yao, and Benjamin~K. Tsou. 2008.
\newblock \href {https://aclanthology.org/C08-1143} {Active learning with sampling by uncertainty and density for word sense disambiguation and text classification}.
\newblock In \emph{Proceedings of the 22nd International Conference on Computational Linguistics (Coling 2008)}, pages 1137--1144, Manchester, UK. Coling 2008 Organizing Committee.

\end{thebibliography}
\bibliographystyle{acl_natbib}

\end{document}